\documentclass{article}

\PassOptionsToPackage{numbers, compress}{natbib}

\usepackage[final]{neurips_2024}

\usepackage{placeins}

\usepackage[utf8]{inputenc} 
\usepackage[T1]{fontenc}    
\usepackage{kotex}
\usepackage{hyperref}       
\usepackage{url}            
\usepackage{booktabs}       
\usepackage{amsfonts}       

\usepackage{nicefrac}       
\usepackage{microtype}      
\usepackage{xcolor}         
\usepackage{wrapfig}
\usepackage{graphicx}

\title{Deep Support Vectors}

\usepackage{amsthm}
\usepackage{amsmath}
\usepackage{amssymb}
\usepackage{mathtools}
\usepackage{comment}
\usepackage{multirow}
\usepackage{graphicx}

\usepackage{bbm}

\usepackage{subcaption} 
\usepackage{algorithm}
\usepackage{algpseudocode}
\usepackage{enumitem}

\author{%
  Junhoo Lee \and Hyunho Lee \and Kyomin Hwang \and Nojun Kwak\thanks{Corresponding Author} \\
  Seoul National University\\
  \texttt{\{mrjunoo, hhlee822, kyomin98, nojunk\}@snu.ac.kr} \\
}

\DeclareMathOperator*{\argmax}{arg\,max} 
\DeclareMathOperator*{\argmin}{arg\,min}

\begin{document}

\maketitle

\newcommand{\hh}[1]{\textcolor{black}{#1}}
\newcommand{\gm}[1]{\textcolor{black}{#1}}
\newcommand{\jh}[1]{\textcolor{black}{#1}}
\newcommand{\nj}[1]{\textcolor{black}{#1}}
\newcommand{\ie}{\textit{i}.\textit{e}., }
\newcommand{\eg}{\textit{e}.\textit{g}., }

\begin{figure}[h] 
\centering 
\includegraphics[width=1.0\textwidth]{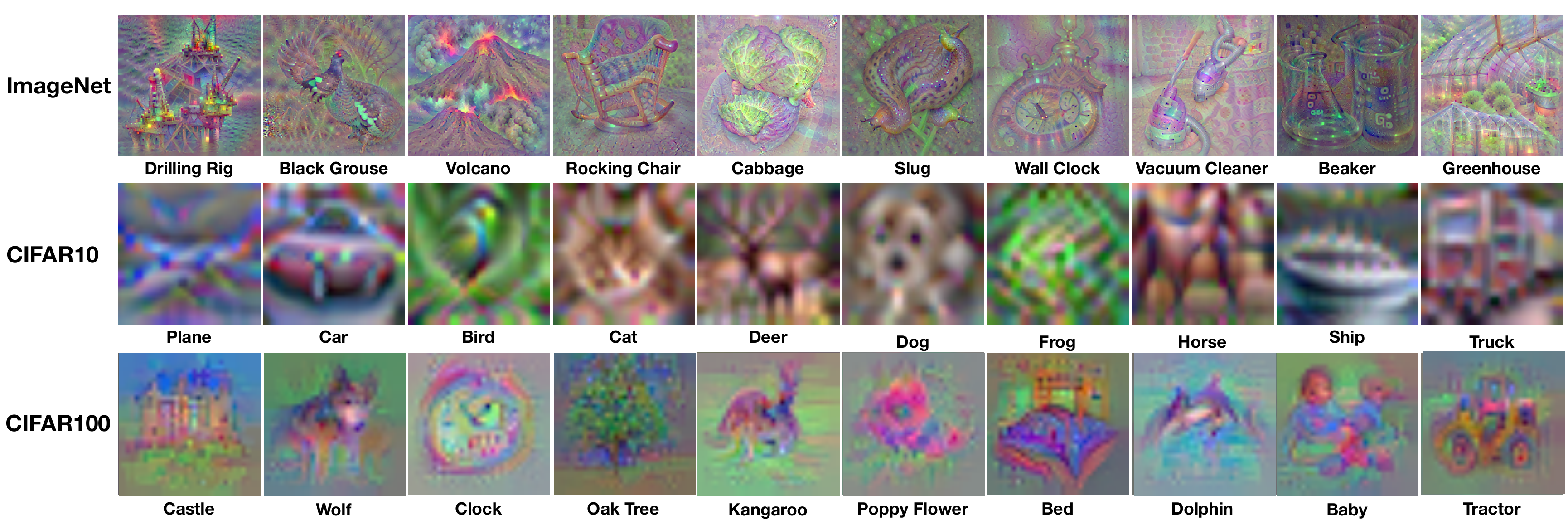}
\caption{\jh{Generated images \nj{using a model trained with} the ImageNet, CIFAR10, and CIFAR100 datasets, \nj{respectively}. \textbf{Each image was generated without \nj{referencing} the original training data.}}}
\label{fig:generated} 
\end{figure}


\begin{abstract}
\jh{Deep learning has achieved tremendous success. \nj{However,} unlike SVMs, which provide direct decision criteria and can be trained with a small dataset, it still has significant weaknesses due to its requirement for massive datasets during training and the black-box characteristics on decision criteria. \nj{This paper addresses} these issues by identifying support vectors in deep learning models. To this end, we propose the DeepKKT condition, an adaptation of the traditional Karush-Kuhn-Tucker (KKT) condition for deep learning models, and confirm that generated Deep Support Vectors (DSVs) using this condition exhibit properties similar to traditional support vectors. This allows us to apply our method to few-shot dataset distillation problems and alleviate the black-box characteristics of deep learning models. Additionally, we demonstrate that the DeepKKT condition can transform conventional classification models into generative models with high fidelity, particularly as latent \jh{generative} models using class labels as latent variables. We validate the effectiveness of DSVs \nj{using common datasets (ImageNet, CIFAR10 \nj{and} CIFAR100) on the general architectures (ResNet and ConvNet)}, proving their practical applicability. (See Fig.~\ref{fig:generated})}

\end{abstract}

\section{Introduction}\label{sec:intro}

Although deep learning has gained enormous success, it requires huge amounts of data for training, and its black-box characteristics regarding decision criteria result in a lack of reliability. For example, CLIP {\cite{CLIP} needs 400 million image pairs for training and Stable Diffusion XL (SDXL) {\cite{SDXL} requires 5 billion images. This implies only a small number of groups can train foundation models from scratch. Also, the black-box nature makes it hard to anticipate the model’s performance in different environments. For example, suppose we are to classify pictures of deer and most training deer images contain antlers. For the test images taken in winter,  
the performance will be worse as deer shed their antlers in winter. 
As modern deep learning models do not provide any decision criterion \ie black box, \jh{we} cannot determine whether the domain of the model has shifted, or if the model is biased in advance, thus cannot anticipate the performance drop in this case.

Interestingly, these problems do not occur in previous state-of-the-art, support vector machines (SVMs), which require substantially less data, enabling almost anyone to train a model from scratch. Also, as it encodes the decision boundary explicitly, SVM can reconstruct the support vectors from the training dataset using the KKT condition. 
Since it 
is a white box, one can anticipate the test’s performance in advance.  In the deer classification example, if the model’s support vectors of deer have prominent antlers, using that SVM is not appropriate for photos taken in winter. 

In this paper, we tackle the natural limitations of deep learning -- the need for large data and black-box characteristics -- by extracting SVM features in deep learning models. In doing so, we introduce the DeepKKT condition for deep models, which corresponds to the KKT condition in traditional SVMs. By either selecting deep support vectors (DSVs) from training data or generating them from already trained deep learning models,
we show DSVs can play a similar role to conventional support vectors. Like support vectors can reconstruct SVM, we can reconstruct the deep models from scratch only with DSVs. Also, we show that DSVs encode the decision criterion visually, providing a global explanation for the trained deep model.
\nj{Expanding beyond conventional support vectors, DSVs suggest that a trained deep classification model can also function as a latent generative model by utilizing logits as latent variables and applying DeepKKT conditions.}

To this end, we generalize the KKT condition and define the DeepKKT condition considering that the data handled by a deep model is high-dimensional and multi-class. 
We demonstrate that the selected data points (selected DSVs) among the training data satisfying the DeepKKT condition are closer to the decision boundary than other training data, as evidenced by comparing entropy values.
Also, we show that the calculated Lagrangian multiplier can reveal the level of uncertainty of the model for the sample in question. 
Additionally, we demonstrate that the  DSVs outperform existing algorithms in the few-shot dataset distillation setting, where only a portion of the training set is used, indicating that DSVs exhibit characteristics similar to SVMs. Moreover, we confirm that modifying existing images using information obtained from DSVs allows us to change their class at will, verifying that DSVs can meaningfully explain the decision boundary. Finally, by using soft labels as latent variables in ImageNet, we generate unseen images with high fidelity.


Our contributions are as follows:

\begin{itemize}[leftmargin=*]
    \item By generalizing the KKT condition for deep models, we propose the DeepKKT condition to extract Deep Support Vectors, which are applicable to general architectures such as ConvNet and ResNet.
    \item We verify that DeepKKT can be used to extract and generate DSVs for common image datasets such as CIFAR10, CIFAR100, SVHN, and ImageNet.
    \item DSVs are better than existing algorithms in few-shot dataset distillation problems. 
    \item \jh{DeepKKT acts as a novel model inversion algorithm that can be applied in practical situations.}
    \item By using the DeepKKT condition, we not only show the trained deep models can reconstruct data, but also can serve as latent generative models by using logits as latent.
\end{itemize}

\section{Related Works}
\label{sec:relwork}



\subsection{SVM in Deep Learning}
Numerous studies have endeavored to establish a connection between deep learning and conventional SVMs. 
\jh{In the theoretical side}, \cite{svm:theory} demonstrated that, in an overparameterized setting with linearly separable data, the learning dynamics of a model possessing a logistic tail are equivalent to those of a support vector machine, with the model’s normalized weights converging towards a finite value. Following this, \cite{svmhomo} extended this equivalence to feedforward networks. \jh{\nj{This} line of research relies on strong assumptions such as full-batch training \nj{and} non-residual architecture \nj{without data augmentation}.}
\jh{There also exists a body of work on integrating SVM principles into deep learning, often referred to as DeepSVM, aiming to leverage \nj{SVM’s desirable} properties} \cite{svmpractical, svmpractical2, SVMex1,SVMex2,SVMEX3}. \jh{DeepSVM integrates SVM components, specifically using them as feature extractors to derive meaningful, human-crafted features.}

In contrast, our work does not modify or incorporate SVM architectures. Instead, we focus on identifying support vectors directly within deep learning models, thereby bridging the gap between deep learning and support vector machines in a more fundamental manner. Despite these advancements, there remains a lack of research that directly connects support vectors through a theoretical lens of equivalence. In this study, we address this gap by introducing the DeepKKT condition, a KKT condition tailored for deep learning, allowing us to apply the concept of support vectors in a practical deep learning context.

\jh{We \nj{show} that reconstructing support vectors in deep models is indeed feasible, and obtaining high-quality support vectors is achievable under much less restrictive conditions compared to prior work.}

\subsection{\jh{Model Inversion Through the Lens of Maximum Margian}}

There is a line of research utilizing the stationarity condition, \nj{a} part of the KKT condition, for model inversion. \cite{svm:marginreconstruct} firstly exploited the KKT condition for model generation, adopting SVM-like architectures. They \jh{normally} conducted experiments with binary classification, a 2-layer MLP, and full-batch gradient descent. \cite{svm:reconstructmulti} extended these experiments to multi-label classification by adapting their existing architecture to a multi-class SVM structure. To ensure the generated samples lie on the data manifold, 
\jh{they initialized with the dataset's mean,}
implying the adoption of some prior knowledge of the data. Similarly, \cite{margin:genfclassifer} generated images through the stationarity condition, also adopting the mean-initialization and conducting experiments on the \nj{CIFAR10 \cite{dataset:cifar10},} MNIST \cite{data:mnist} and \jh{downsampled} CelebA~\cite{dataset:celebA} datasets. 
\nj{Their work generally focused on low-dimensional, labeled datasets with a small number of classes such as CIFAR10 and MNIST, consistent with the traditional SVM setting of binary-labeled, low-dimensional datasets. In contrast, we extended our experiments to high-dimensional datasets with many classes, an area traditionally dominated by deep learning. Specifically, we conducted experiments on ImageNet~\cite{imagenet} using a pretrained ResNet50 model following the settings described in the original paper \cite{resnet}.}


\nj{Furthermore, previous works have concentrated on \textbf{reconstructing the training} dataset. In contrast, similar to generative models, our work focuses on \textbf{generating unseen} data from noise using a classification model. Additionally, we emphasize the original meaning of \hh{`support vectors'.} Unlike other approaches, our Deep Support Vectors (DSVs) adhere to the traditional role of support vectors: they explain the decision criteria, and a small number of DSVs can effectively reconstruct the model.}

\subsection{Dataset Distillation}
Dataset distillation \cite{distillation_trajectory,distillation:grad1,distillation:nongrad1} fundamentally aims to reduce the size of the original dataset while maintaining model performance. 
\jh{The achievement also involves addressing privacy concerns and alleviating communication issues between \nj{the} server \nj{and client} nodes. }
The \nj{dataset distillation} problem is typically addressed under the following conditions: 1) Access to the entire dataset for gradient matching, 2) Possession of snapshots from all stages of the model's training phase, which \nj{are impractical settings} \nj{for practical usage} \cite{practicalDD}.  \jh{Furthermore, these algorithms typically }\hh{require Hessian computation, which imposes a heavy computational burden.}

\jh{In SVM, the model can be reconstructed using support vectors. This reconstruction is more practical compared to previous dataset distillation methods, as it does not require any of the restrictive conditions. }
\jh{Likewise, because Deep Support Vectors (DSVs) also do not require these conditions and \nj{are Hessian-free}, they can play the role of distillation under practical conditions.}

\section{Preliminaries}\label{sec:prelim}



\subsection{Notation} \label{sec:notation}

In \nj{the} SVM formulation, \( \tilde{w} (:= [w;b])\) represents the concatenated weight vector $w$ and bias $b$. 
Each data instance, expanded to include the bias term, is denoted by \( \tilde{x}_i (:= [x_i;1])\), while the corresponding binary label is represented by \( y_i \in \{\pm1\}\). The Lagrange multipliers are denoted by \( \alpha_i \)'s. 

Transitioning to the context of deep learning, we denote the parameter vector of a neural network model by \( \theta \), which, upon optimization, yields \( \theta^* \) as the set of learned weights. The mapping function \( \Phi(x_i; \theta) \) represents the transformation of input data into 
a $C$-dimensional logit in a $C$-class classification problem in a manner dictated by the parameters \( \theta \), \ie $\Phi(x_i; \theta) = [\Phi_1(x_i; \theta), \cdots, \Phi_C(x_i; \theta)]^T \in \mathbb{R}^C$. 
\nj{We} define \nj{the} \textbf{score} as the logit of a target \nj{class, \ie} $\Phi_{y_i}(x_i; \theta)$. 
If the score is the largest among logits, \ie $\argmax_c \ \Phi_{c}(x_i; \theta) = y_i$, then it correctly classifies the sample. 
The Lagrange multipliers adapted to the optimization in deep learning are represented by \( \lambda_i \)'s. $(x, y)$ denotes a pair of input and output and \( \mathcal{I} \) is the index set \nj{with $|\mathcal{I}| = n$.}

\subsection{Support Vector Machines}
\label{sec:KKTcondition}

The fundamental concept of Support Vector Machines (SVMs) is to find the optimal hyperplane that classifies the given data. This hyperplane is defined by the closest data points to itself known as support vectors, and the distance between the support vectors and the hyperplane is termed the margin. The hyperplane must classify the classes correctly while maximizing the margin. This leads to the following \jh{KKT} conditions that an SVM  must satisfy: \nj{
\textbf{(1) Primal feasibility:} $\forall i,\ y_i \tilde{w}^T \tilde{x}_i \geq 1$,  \textbf{(2) Dual feasibility:} $\forall i,\ \alpha_i \geq 0$, \textbf{(3) Complementary slackness:}  $\alpha_i \left( y_i \tilde{w}^T \tilde{x}_i - 1 \right) = 0$ and \textbf{(4) Stationarity:} $\tilde{w} =   \sum_{i=1}^n \alpha_i y_i \tilde{x}_i$.}
%
%

The primal and dual conditions ensure these critical values correctly classify the data while being outside the margin. The complementary slackness condition mandates that support vectors lie on the decision boundary. 
The stationarity condition \nj{ensures} that the derivative of the Lagrangian is zero. 

The final condition, stationarity, offers profound insights into \nj{SVMs}. It underscores that support vectors \nj{encode} the decision boundary $\tilde{w}$. Consequently, identifying the decision hyperplane in SVMs is tantamount to pinpointing the corresponding support vectors. This implies that with a trained model at our disposal, we can reconstruct the SVM in two distinct ways:

1. \textbf{Support Vector Selection:} From the trained model, we can extract support vectors \nj{among the training data} that inherently encode the decision hyperplane.

2. \textbf{Support Vector Synthesis:} Alternatively, it is feasible to generate or synthesize support vectors, even in the absence of a training set, which can effectively represent the decision hyperplane by generating samples that satisfy $|\tilde{w}^T\tilde{x}| = 1$.

\section{Deep Support Vector}\label{sec:DSV}


This section presents the specific conditions that DSVs (Deep Support Vectors) must satisfy and discusses how to get an optimization loss \jh{to meet these conditions.}

\subsection{DeepKKT}
\label{sec:deepkkt}

\paragraph{
SVM's Relationship with Hinge loss}

We start our discussion by focusing on the \textit{hinge loss,} a continuous surrogate loss for the primal feasibility, and its gradient:
\begin{equation}
\begin{split}
&\text{Hinge Loss:} \ L_h(x_i, y_i; \tilde{w}) = \frac{1}{n} \sum_{i=1}^n \max(0, 1 - y_i(\tilde{w}^T \tilde{x}_i)), \\
\label{eq:hingegrad}
& \quad\nabla_{\tilde{w}}  L_h = \frac{1}{n} \sum_{i=1}^n \begin{cases} 0, & \text{if } y_i(\tilde{w}^T \tilde{x}_i) \ge 1 \\ -y_i \tilde{x}_i, & \text{otherwise.} \end{cases}
\end{split}
\end{equation}
With Eq.~(\ref{eq:hingegrad}), the stationarity condition \nj{of SVM} becomes 
\begin{equation}\label{eq:hingesvm}
w^* := -\sum_{i=1}^{n} \alpha_i \nabla_w L_{\jh{h}}(x_i, y_i;w^*), \quad \text{s.t.} \  \alpha_i \ge 0.
\end{equation}

\paragraph{Generalization of conventional KKT conditions}

In this paper, we \nj{extend} the KKT conditions 
to deep learning. In doing so, the two main hurdles of a deep network different from a linear binary SVM are 1) the nonlinearity of $\Phi(x_i;\theta)$ taking the role of $\tilde{w}^T \tilde{x}_i$ and 2) multi-class nature of a deep learning model.

Considering \nj{that} the role of the \nj{primal feasibility} condition is to correctly classify $x_i$ into $y_i$, we can enforce the score $\Phi_{y_i}(x_i;\theta)$ for the correct class $y_i$ to take the maximum value among all the logits with some margin $\epsilon$, \ie 
\begin{equation}
    \Phi_{y_i}(x_i; \theta^*) - \max_{c\neq y_i, c\in [C]} \Phi_{c}(x_i; \theta^*) \ge \epsilon, 
    \label{eq:primal}
\end{equation} 
\jh{We can relax \nj{this discontinuity with a continuous surrogate function} $-L(\Phi(x_i; \theta^*), y_i)$, which is the negative loss function to maximize. Note that if we take the cross-entropy loss for $L$, it becomes}
\begin{equation}
-L_{ce} = \Phi_{y_i}(x_i; \theta^*) - \log \sum_{c=1}^C \exp (\Phi_{c}(x_i; \theta^*)),
\label{eq:primal_ce}
\end{equation}
which takes a similar form as Eq.~(\ref{eq:primal}) and the negative loss can be maximized to meet the condition.

\begin{table}[t]
  \centering
  \resizebox{0.5\linewidth}{!}{
    \begin{tabular}{l|l|rr}
      \hline
      img/cls & ratio (\%) & \multicolumn{1}{l}{Random} & \multicolumn{1}{l}{selected DSVs } \\ \hline
      50      & 1          & 46.16 $\pm$ 1.93           & 48.91 $\pm$ 0.90               \\
      10      & 0.2        & 30.08 $\pm$ 1.96           & 33.69 $\pm$ 2.05               \\
      1       & 0.02       & 14.26 $\pm$ 0.99           & 16.83 $\pm$ 0.29               \\ \hline
    \end{tabular}
  }
  \vspace{5mm}
  \caption{\jh{In coreset selection benchmarks \nj{using the CIFAR-10 dataset, the DeepKKT condition is used as the selection criterion. Images with the highest $\lambda$ for each class were chosen to train a network. }}}
  \label{table:coreset}
\end{table}

Now that we found the analogy between $y_i\tilde{w}^T\tilde{x}_i$ and $-L(\Phi(x_i; \theta^*), y_i)$, $y_i \tilde{x}_i (=\nabla_{\tilde{w}} (y_i\tilde{w}^T\tilde{x}_i))$ corresponds to $-\nabla_{\theta^*} L(\Phi(x_i; \theta^*), y_i)$. Thus, the stationary condition $\tilde{w} = \sum_{i=1}^n \alpha_i y_i \tilde{x}_i$ in SVMs can be translated into that of deep networks such that
\begin{equation}
\theta^* = - \sum_{i=1}^n \lambda_i \nabla_{\theta^*} L(\Phi(x_i; \theta^*), y_i).
\end{equation}
This condition is a generalized formulation of Eq.~(\ref{eq:hingesvm}), where we substitute the linear model $\tilde{w}^T \tilde{x}$ with a nonlinear model $\Phi(x; \theta)$, and the binary classification hinge loss $L_h$ with multi-class classification loss $L$. \jh{Furthermore, the stationarity condition reflects the dynamics of overparameterized deep learning models. We provide an analogy \nj{with respect to \cite{svm:theory}} in Appendix~\ref{sec:overparam}.}

However, these conditions are not enough for deep learning. As mentioned before, \jh{we are interested in dealing with high dimensional manifolds.} Compared to the problems dealt with in the classical SVMs, the input dimensions of deep learning problems are typically much higher. 
In this case, the data are likely to lie along a low-dimensional latent manifold $\mathcal{M}$ inside the high-dimensional space. 
To \nj{make} a generated sample be a plausible DSV,  it not only should satisfy the generalized KKT condition but also should lie in an appropriate data manifold, \ie $x \in \mathcal{M}$.

Finally, we can rewrite the new \textbf{DeepKKT condition} as follows:
\begin{equation} \label{eq:DeepKKT}
\begin{aligned}
&\text{Primal feasibility:} && 
\forall\, i \in \mathcal{I}, \quad \argmax_c\; \Phi_c (x_i; \theta^*) = y_i \\
&\text{Dual feasibility:} && \forall\, i \in \mathcal{I}, \quad \lambda_i \geq 0, \\
&\text{Stationarity:} && \theta^* = -\sum_{i=1}^n \lambda_i \nabla_{\theta} \mathcal{L}(\Phi(x_i; \theta^*), y_i), \\
&\text{Manifold:} && \forall\, i \in \mathcal{I}, \quad x_i \in \mathcal{M}. 
\end{aligned}
\end{equation}
\subsection{\jh{Deep Support Vectors, From Dust to Diamonds}}
\label{sec:DKKTcondition}

\nj{Sec.~\ref{sec:deepkkt} explored} the KKT conditions in the context of deep learning and how these conditions can be used to formulate a loss function. It is important to note that our goal is not to construct the model $\Phi$, but rather to generate \textbf{support vectors of an already-trained model $\Phi$ with its parameter $\theta^*$}.

\nj{To reconstruct support vectors from a trained deep learning model, sampling or synthesizing support vectors is essential}.
By replacing the optimization variable $\theta$ with $x$, we shift our focus to utilizing the DeepKKT conditions for generating or evaluating input $x$ rather than $\theta$. This adjustment necessitates a consideration of the data's inherent characteristics, specifically its multiclass nature and the tendency of the data to reside on a lower-dimensional manifold \nj{in an ambient space.} 

\paragraph{Primal Feasibility}
Firstly, the primal feasibility condition \nj{in Eq.~(\ref{eq:DeepKKT}) mandates support vectors be correctly classified by the trained model $\theta^*$.}  
As presented in Sec.~\ref{sec:deepkkt}, instead of Eq.~(\ref{eq:primal}), we use a surrogate function 
for the loss: 
\begin{equation}\label{eq:two_step_loss}
L_{\text{primal}} = \frac{1}{n} \sum_{i=1}^n L_i, \quad \text{where} \quad
L_i = 
\begin{cases} 
0 & \text{if } \argmax_c \Phi_c(x_i; \theta^*) = y_i, \\
L(\Phi(x_i; \theta^*), y_i) & \text{otherwise}.
\end{cases}
\end{equation}
%
This is designed to match the \nj{primal} condition by mimicking the Hinge loss. When each DSV is correctly classified, no loss is incurred. Otherwise, we adjust the DSVs to align them with the correct target, effectively optimizing the support vectors. This approach also implicitly enforces the complementary slackness condition, as $L_i$ \nj{decreases} confidence in the incorrect classification. 
\begin{figure}[t]  
  \centering
  \resizebox{0.8\linewidth}{!}{%
    \begin{subfigure}[b]{0.47\linewidth}
      \includegraphics[width=\linewidth]{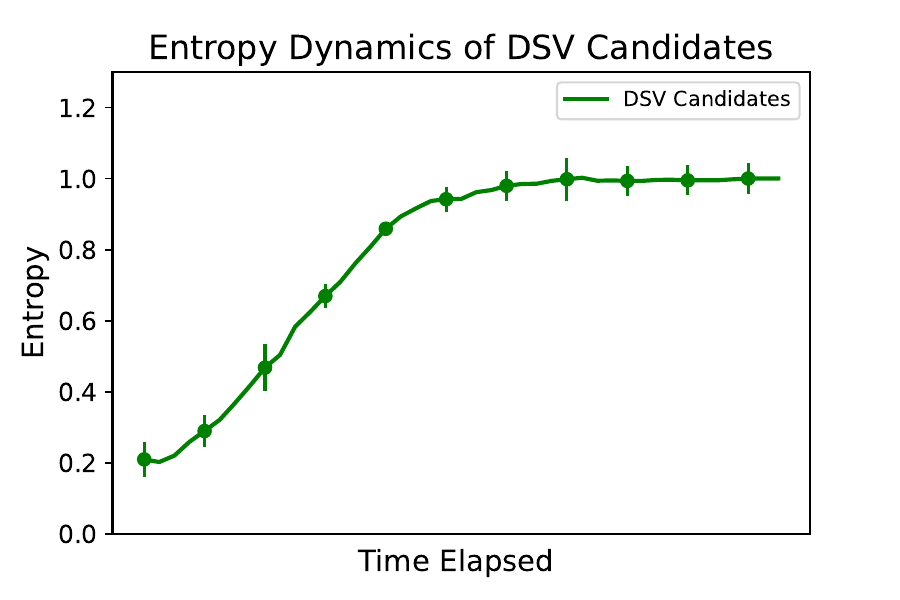}
      \phantomcaption
      \label{fig:lambda_entropy}
    \end{subfigure}
    \hfill
    \begin{subfigure}[b]{0.53\linewidth}
      \includegraphics[width=\linewidth]{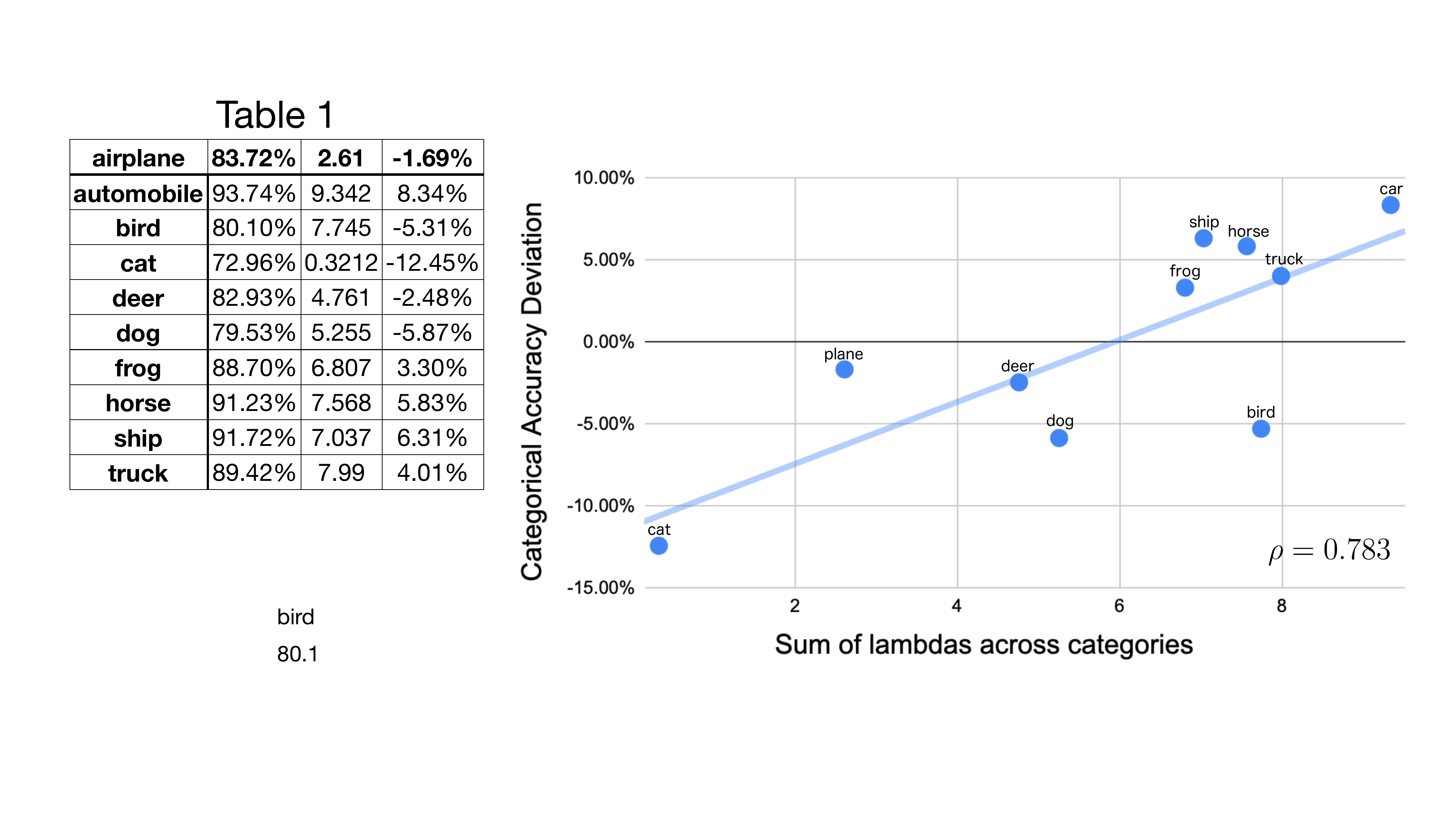}
      \phantomcaption
      \label{fig:correlation}
    \end{subfigure}
  }
  \caption{Characteristics of DSVs; (Left) Entropy change of DSV candidates over time, (Right) Correlation between classwise mean test accuracy and the sum of $\lambda$'s}
  \label{fig:DSV_characteristics}
\end{figure}


Here, $L$ can be any loss function 
and we have employed the cross-entropy loss in our experiments.

\paragraph{Stationarity}
Secondly, the stationarity condition can be used directly as a loss function. Since we are extracting DSVs from the trained model $\Phi(\cdot ; \theta^*)$, 
we construct this loss as follows:
\begin{equation}\label{eq:stationarity}
\hh{L_{\text{stat}} = D(\theta^*, -\sum_{i=1}^n \lambda_i \nabla_\theta L(\Phi(x_i;\theta^*), y_i))}.
\end{equation}
%
%
%
%
%
%
%
%
For the distance measure $D$, any metric can be used; we have chosen to use the $l_1$ distance to suppress the effect of outliers. It is crucial to remember that our objective is to find DSVs and the optimization is done for the primal and dual variables $x_i$ and $\lambda_i$ and not for the parameter $\theta$. 
For this, we require one forward pass and two backward passes; one for $\nabla_\theta L$ and the other for $\nabla_{x_i} D$. The overall computational cost is quite low, as we optimize only a small number of samples.

Moreover, as shown in \nj{Algorithm~\ref{alg:KKT} (Appendix \ref{sec:alg})}, 
we satisfy the dual condition by ensuring the Lagrange multipliers $\lambda_i$'s are greater than zero and disqualify any $x_i$'s from being a support vector candidate if during optimization $\lambda_i$ becomes less than zero. The condition that is not \nj{explicitly} satisfied is the complementary slackness. To directly fulfill the functional boundary for support vectors as specified in Sec.~\ref{sec:KKTcondition}, we would need to be able to calculate the distance between functions, which is not only abstract but also requires a second-order computation cost. Therefore, we adopted a relaxed version of the KKT conditions that excludes this requirement. Furthermore, as demonstrated in Sec.~\ref{sec:slackness}, we have shown that DSVs implicitly satisfy the complementary slackness condition. This implies that DSVs meet every condition introduced in conventional SVM.



\paragraph{Manifold Condition: Reflecting High-Dimensional Dynamics of Deep Learning}

As modern deep learning deals with extremely high-dimensional spaces, imposing additional constraints other than the primal feasibility and stationarity conditions is needed so that DSVs reside in the desired data manifold. To achieve this, we add a manifold condition, which enforces that the DSVs lie on the data manifold. By selecting DSVs that are in the intersection of the solution subspace and the data manifold, we can properly represent both the model and the training dataset.


To extract DSVs from the manifold, we assume that the model is well-trained, meaning it maintains consistent decisions despite data augmentation. In other words, the model should classify DSVs invariantly even after augmentation. To ensure this, we enforce that the augmented DSVs ($\mathcal{A}(x)$ \jh{where $\mathcal{A}$ denotes augmentation function}) also meet the primary and stationarity conditions.


Also, we exploit traditional image prior \cite{total_prior, total_prior1}, total variance $L_\text{tot}$ and size of \nj{the} norm $L_{norm}$ \nj{to} make DSVs lie in \nj{the data} manifold. $L_\text{tot}$ is calculated by summing the differences in brightness between neighboring pixels, \nj{reducing} unnecessary noise in an image, and \nj{maintaining} a natural appearance. $L_{norm}$, \nj{taking a similar role,} penalizes the outlier and preserves important pixels.

\nj{Finally our DSV is obtained as follows where $\mathbb{E}_\mathcal{A}$ represents expectation over augmentations:}
\begin{equation} \label{eq:main_equation}
\text{DSV} = \argmin_x \mathbb{E}_\mathcal{A}  \left[L_{\text{stationarity}}(\mathcal{A}(x)) + \beta_1 L_{\text{primal}}(\mathcal{A}(x)) +  \beta_2 L_{\text{tot}}(x) + \beta_3 L_{\text{norm}}(x)\right].
\end{equation}


\nj{One might wonder if there is a better sampling strategy than using DSVs, such as sampling far from the decision boundary instead of near it. We argue that in a high-dimensional data manifold, most data points are located close to the decision boundary because, in a data-scarce, high-dimensional space, every sample matters and thus would serve as a DSV.}

\section{Experiments}\label{sec:CDSV}

\subsection{\gm{DSVs: Revival of Support Vectors in Deep Learning}}
\begin{wrapfigure}{t}{5.2cm}
  \centering
   \includegraphics[width=5cm]{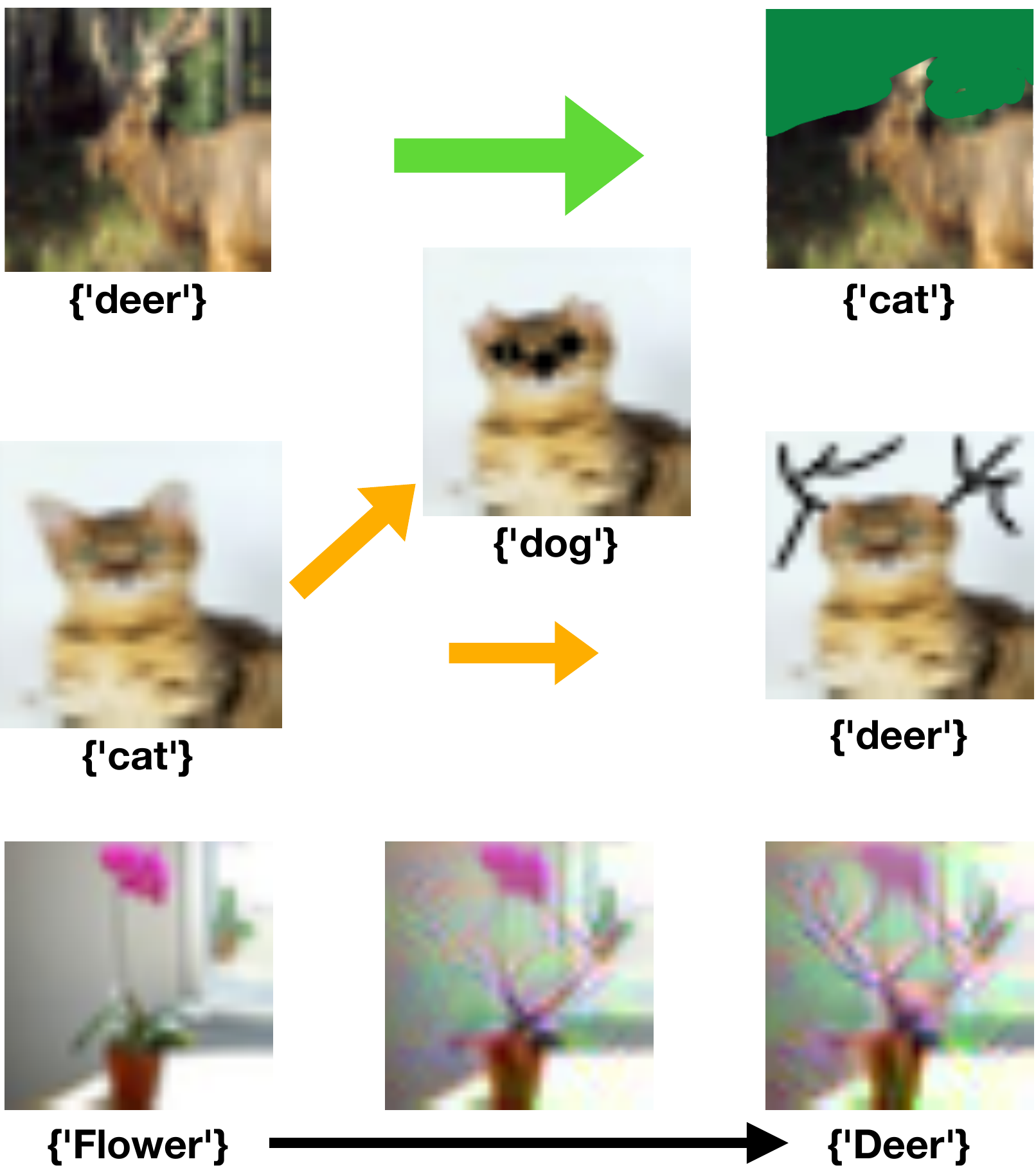}
   \caption{\jh{Model predictions for original versus DSV-informed edited images. (Top) Images were altered manually \nj{based on} decision criteria derived from DSVs, influencing \nj{the model's prediction. (Bottom) Images were altered based on DeepKKT loss.}} 
   }
   \label{fig:Painting}
\end{wrapfigure}


\paragraph{DSVs meet SVM characteristics}\label{sec:slackness}

\jh{As discussed in Section~\ref{sec:KKTcondition}, the principle of complementary slackness within the KKT conditions suggests that support vectors should be situated on the decision boundary, implying that support vectors typically exhibit high uncertainty from a probabilistic perspective, \ie they possess high entropy. While DeepKKT does not explicitly incorporate the \nj{complementary} slackness condition due to computational costs and ambiguity, \nj{Fig.~\ref{fig:lambda_entropy} suggests} that DSVs implicitly fulfill this condition; During the training process, we observe an increase in the entropy of DSV candidates, \nj{hinting that the generated DSVs are close to the decision boundary}.}

\jh{\nj{In addition, we can infer the importance of a sample in the decision process by utilizing the DeepKKT condition. We} trained the Lagrangian multiplier $\lambda$ \nj{for} each \nj{test} image. Figure~\ref{fig:correlation} shows a strong correlation between the sum of $\lambda$ values for each class and its test accuracy. This finding is intriguing because, despite the model achieving nearly 100\% accuracy during training due to overparameterization, DSVs provide insights into categorical generalization in the test phase. Not only does measuring its credibility indicate that a large $\lambda$ refers to an `important' image for training, but $\lambda$ could also serve as a natural core-set selection measure. Table~\ref{table:coreset} shows this to be true. On the CIFAR-10~\cite{dataset:cifar10} dataset, we selected images with high $\lambda$ values and retrained \nj{the network} with the selected images. In this case, the \nj{selected DSVs show higher test acccuracies} compared to random selection. This characteristic resembles \nj{that of} support vectors, as \nj{the model can be reconstructed} with support vectors.}

Finally, Fig.~\ref{fig:generated} demonstrates the high fidelity of the generated DSVs, providing practical evidence that these DSVs lie on the data manifold.
Similar to how support vectors are reconstructed in \nj{an} SVM, the DeepKKT condition \nj{enables} the reconstruction of these vectors without \nj{referencing} training data. This \nj{shows} the effectiveness and adaptability of \nj{our DeepKKT} in capturing key data features.

\paragraph{DSVs \nj{for} Few Shot Dataset Distillation}
\jh{DeepKKT emerges as a pioneering algorithm tailored for practical \nj{dataset distillation}. \nj{DSVs} addresses two critical concerns: 1) Protecting private information through data synthesis, and 2) Reducing the communication load by minimizing the size of \nj{data transmission}. Traditional distillation algorithms encounter a fundamental paradox; \nj{as} data predominantly originate from edge devices like smartphones \cite{FL1, fl_start, fl2}, \nj{the} requirement to access \nj{the} entire dataset introduces significant communication overhead and heightens privacy concerns.}


%
%

%
%

\nj{Our DeepKKT relies solely on a pre-trained model without relying on the training dataset}. This unique approach eliminates the need for edge devices to store or process large volumes of private data. 
\jh{\nj{As shown in Table~\ref{table:fewshot}, while} traditional methods suffer significant performance drops under these scenarios \hh{and are incapable of implementing zero-shot scenarios. Conversely DeepKKT remains effective, requiring only minimal data: a single image per sample (\ie initialization with real data), or in some cases, no images at all.} For the zero-image setting, we initialized the images with data from other datasets to ensure diversity.}




\paragraph{DSVs Encode the Decision Criteria Visually}

\jh{Our findings suggest that DSVs  not only satisfy the conditions of classical support vectors but also offers a global explanation of \nj{visual} information. }
Fig.~\ref{fig:Painting} experimentally \nj{verifies} our claims and illustrates the practical use of DSVs, \nj{\eg analysis} of \nj{Fig.~\ref{fig:generated}-Cifar10} reveals the decision criteria for classifying deer, cats, and dogs: 1) DSVs highlight antlers in deer, signifying them as a distinctive characteristic. 2) \nj{Pointed triangular ears are a recurring feature in DSVs of cats}. 3) For dogs, a trio of facial dots holds significant importance. Using these observations, we altered a deer's image by erasing its antlers \nj{and reshaping its ears to a pointed contour, which reduced the model's confidence in classifying it as a deer, and caused the model to misclassify it as a cat.} Similarly, by smoothing the ears of a cat image to diminish its classification confidence and then adding antlers or three facial dots, we influenced the model to reclassify the image as a deer or a dog, respectively. 
\jh{Additionally, the DeepKKT-altering case in Fig.~\ref{fig:Painting}\nj{-Bottom} supports our assertions. Altering a flower image to resemble a deer class \nj{by changing the target class in the primal \jh{and dual }feasibility loss}, antlers grew similar to our manipulation.}

\jh{This discovery holds significant implications about making models responsible; it introduces a qualitative aspect to assessing model performance. Consider a deer classification problem again. 
The model in our study would be less suitable, as evidenced by \nj{Fig.~\ref{fig:generated}-Cifar10-deer}, which indicates the model's reliance on antlers for identifying deer -- a feature not present in winter. This shows that DSVs enable us to conduct causal predictions by qualitatively analyzing models, as SVM does. }
\begin{table}[t]
\centering
\resizebox{12cm}{!}{
\begin{tabular}{c|c|c|ccc|c}
img/cls             & shot/class & ratio (\%) & DC~\cite{DC}            & DSA~\cite{DSA}              & DM~\cite{DM}                & DSV\jh{s} \\ \hline
\multirow{5}{*}{1}   
                    & 0          & 0      & -   & - & - &21.68 $\pm$ 0.80   \\
                    & 1          & 0.02      & 16.48$\pm$0.81 & 15.41$\pm$1.91   & 13.03$\pm$0.15 & \textbf{22.69} $\pm$ 0.38   \\
                    & 10         & 0.2   & 19.66$\pm$0.78 & 21.15$\pm$0.58   & 22.42$\pm$0.43 & -   \\
                    & 50         & 1     & 25.90$\pm$0.62 & 26.01$\pm$0.70   & 24.42$\pm$0.29 & -   \\
                    & 500        & 10    & 28.06$\pm$0.61 & 28.20$\pm$0.63   & 25.06$\pm$1.20 & -   \\ \hline
\multirow{4}{*}{10} 
                    & 0          & 0      & -   & - & - & 30.35 $\pm$ 0.99   \\
                    & 10         &    0.2   & 25.06$\pm$1.20 & 26.67$\pm$1.04   & 29.77$\pm$0.66    & \textbf{37.90} $\pm$ 1.69   \\
                    & 50         &   1    & 36.44$\pm$0.52 & 36.63$\pm$0.52   & 36.63$\pm$0.52    & -   \\
                    & 500        & 10    & 43.55$\pm$0.50 & 44.66$\pm$0.59   & 47.96$\pm$0.95 & -   \\ \hline
\multirow{3}{*}{50} 
                    & 0         & 0    & -              & -   & -    & 39.35 $\pm$  0.54   \\
                    & 50         & 1    & 41.22$\pm$0.90              & 41.29$\pm$0.45   & 48.93$\pm$0.92    & \textbf{53.56} $\pm$  0.73   \\
                    & 500        & 10    & 52.00$\pm$0.59              & 52.19$\pm$0.53 & 60.59$\pm$0.41  & -   \\ 
\end{tabular}
}
\vspace{1mm}
\caption{Performance of Few-shot learning on CIFAR10. \nj{`img/cls' and `shot/class' refer to the per-class number of generated images and the training samples used in generating the distilled dataset, respectively. `ratio' is the ratio of the seen samples among the entire training samples. } \jh{0 shot refers \hh{to the distillation task performed without any access to the training data.}}}
\label{table:fewshot}
\end{table}

\subsection{Unlocking the potential of classifier as generator with DeepKKT}

\paragraph{\jh{Practical Model Inversion with DeepKKT}}

\begin{wrapfigure}{r}{6cm}
  \centering
  \vspace{-3mm}
   \includegraphics[width=6cm]{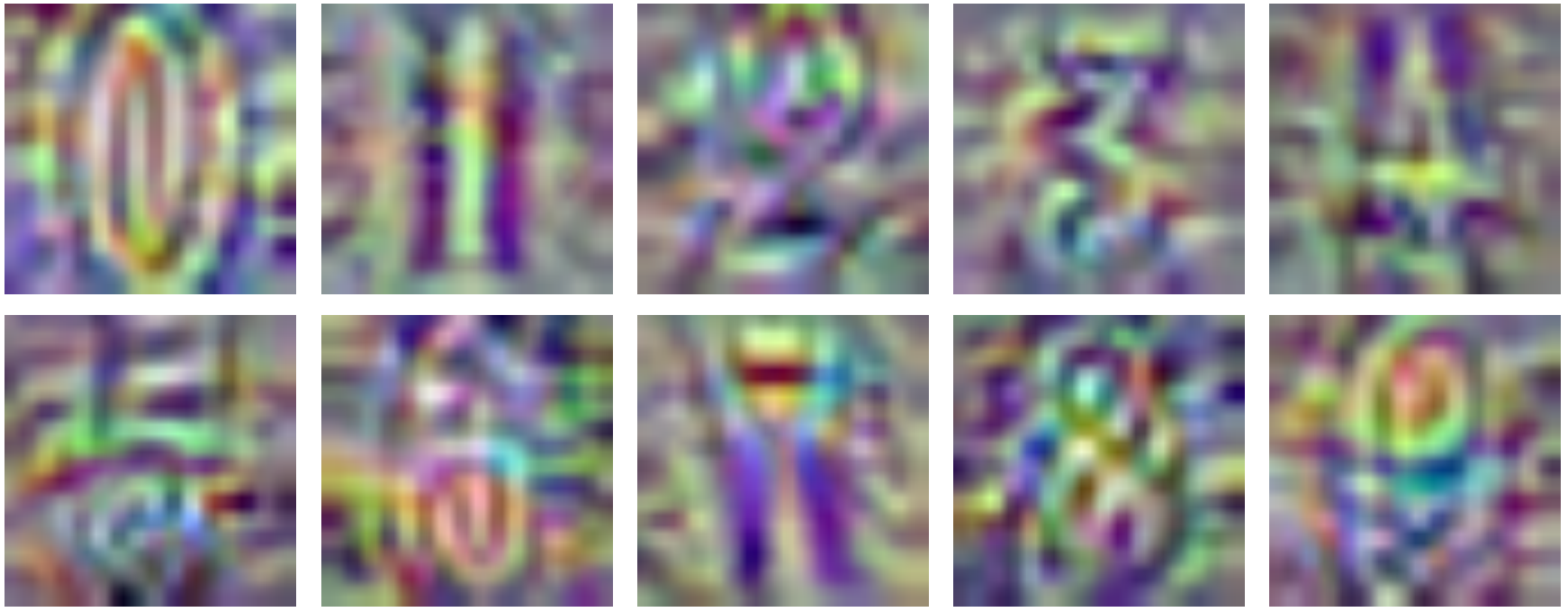}
   \caption{  \jh{DSVs generated by a model that underwent transfer learning from CIFAR-10 to SVHN. During transfer learning, only the last layer is updated \nj{by SVHN}.}}
   \label{fig:transfer}
\end{wrapfigure}

\jh{In cloud environments or APIs, models are deployed with the belief that although they are sometimes trained with sensitive information, their black-box nature prevents users from inferring the data. This belief makes it possible to deploy sensitive models.}
However, as demonstrated in Fig~\ref{fig:generated}, this belief is no longer valid. Fig.~\ref{fig:transfer} further illustrates that model inversion remains feasible even in practical scenarios \nj{such as} transfer learning scenarios, where only specific layers of a foundation model are fine-tuned. Remarkably, DeepKKT conditions enable model inversion even in these challenging environments, suggesting that they can be applied to a subset of the parameter space rather than the entire parameter space \ie \nj{a} more relaxed condition.



\paragraph{Classifier as Latent generative model}

\jh{Considering the impressive capabilities of DSVs in the \nj{image} generation domain, and the geometric interpretation that DSVs \nj{are samples near the decision} boundaries, 
it is \nj{noted} that \nj{enforcing the DeepKKT condition} resembles the \nj{diffusion process.} In each iteration, the DSV develops through the DeepKKT condition as follows: 
}
\begin{equation}
x_{t+1} = x_{t} - \nj{\eta} \cdot \left( [ \nabla_x L_{\text{stat}}(\mathcal{A}(x_t)) +  \beta_2 L_{\text{tot}}(x_t) + \beta_3 L_{\text{norm}}(x_t) ]  + \beta_1 \nabla_x L_{\text{primal}}(\mathcal{A}(x_t)) \right).
\label{eq:update}
\end{equation}
\nj{This is similar to the generalized form} of the \nj{score-based} diffusion process:
\begin{equation}
x_{t+1} = x_t + \epsilon_t \cdot \left( \nabla_x \log p(x_t) + \gamma \nabla_x \log p(y | x_t) \right).
\end{equation}

\begin{wrapfigure}{r}{6cm}
  \centering
   \includegraphics[width=\linewidth]{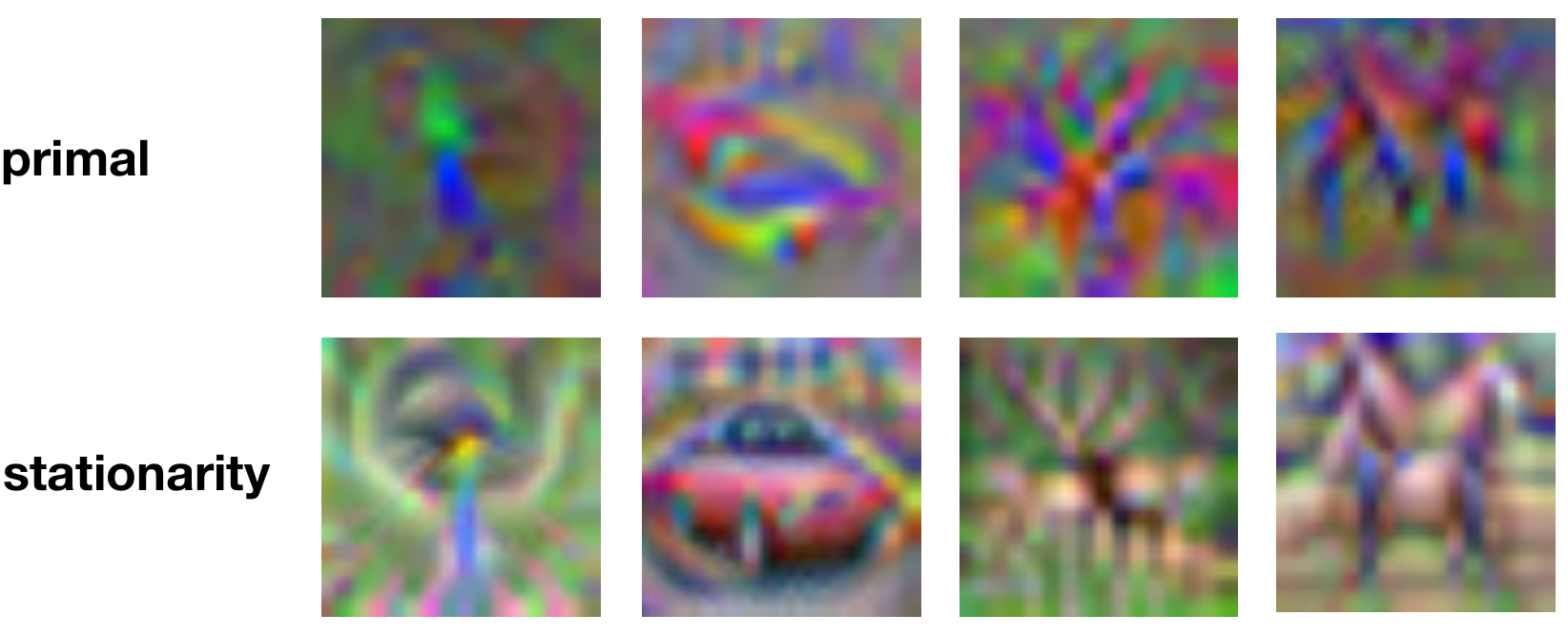}
   \caption{Results showing \jh{DeepKKT} images created solely by the primal condition or by the stationary condition. \jh{\nj{A sole usage} of the primal condition shows low fidelity.}}
   \label{fig:primalstat}
   \vspace{-5mm}
\end{wrapfigure}
\jh{The first three loss terms in Eq.~(\ref{eq:update}) aim to maximize the score ($\nabla_x \log p(x_t)$), while the last term, the primal feasibility term, corresponds to the guidance term ($\gamma \nabla_x \log p(y | x_t)$). As shown in Fig.~\ref{fig:primalstat}, when only the primal loss term is used, meaningful DSV samples are not generated. This indicates that the other losses (stationarity and manifold terms) function update the image towards manifold \ie score function.
From this perspective, an arbitrarily assigned label $y$ can be used as a latent variable for guidance.}


\begin{figure}[t]
  \centering
  \begin{subfigure}[b]{\linewidth}
    \centering
    \includegraphics[width=0.8\linewidth]{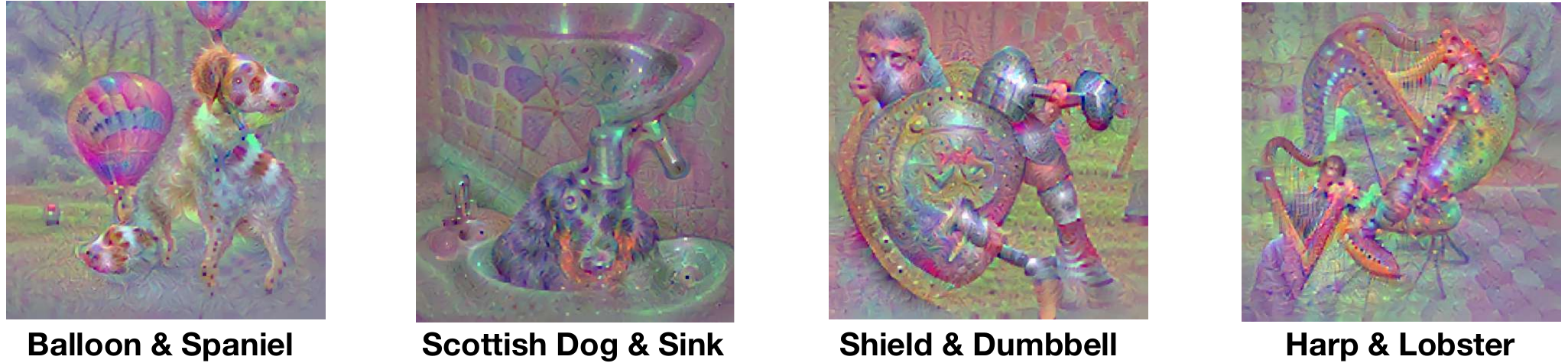}
    \label{fig:latent1}
  \end{subfigure}
  \begin{subfigure}[b]{\linewidth}
    \centering
    \includegraphics[width=\linewidth]{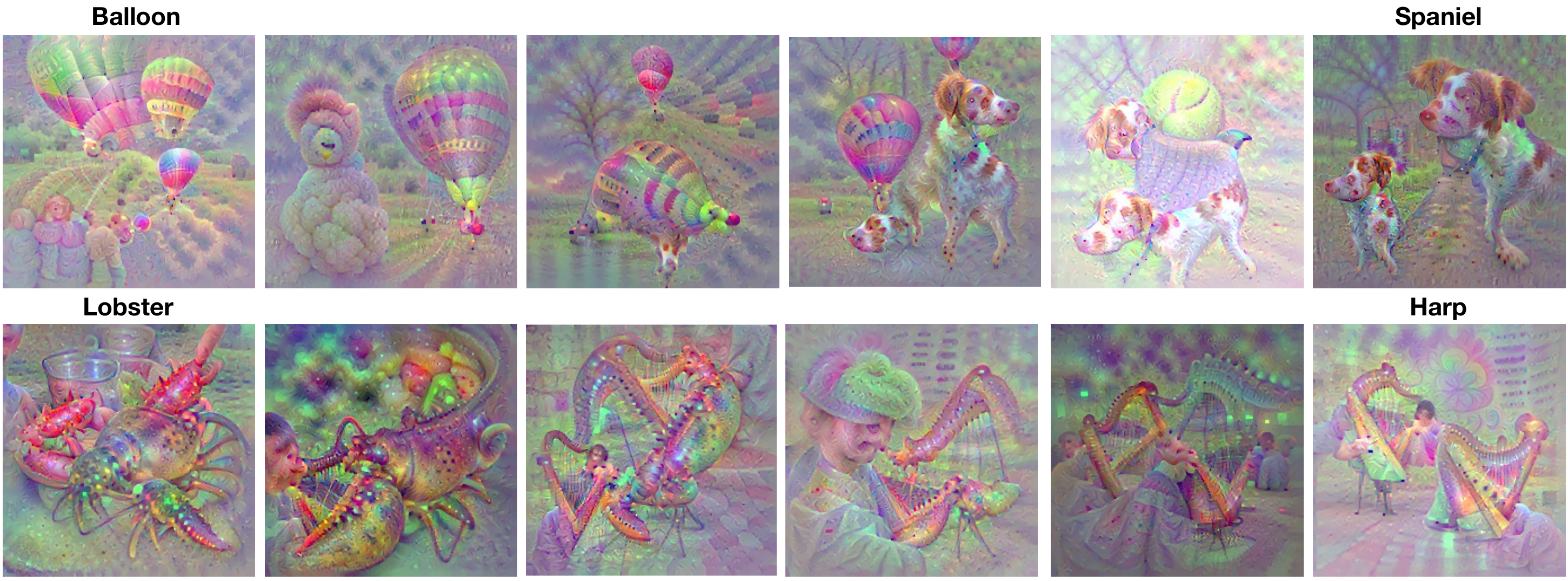}
    \label{fig:latent2}
  \end{subfigure}  
  \begin{subfigure}[b]{\linewidth}
      \centering
    \includegraphics[width=\linewidth]{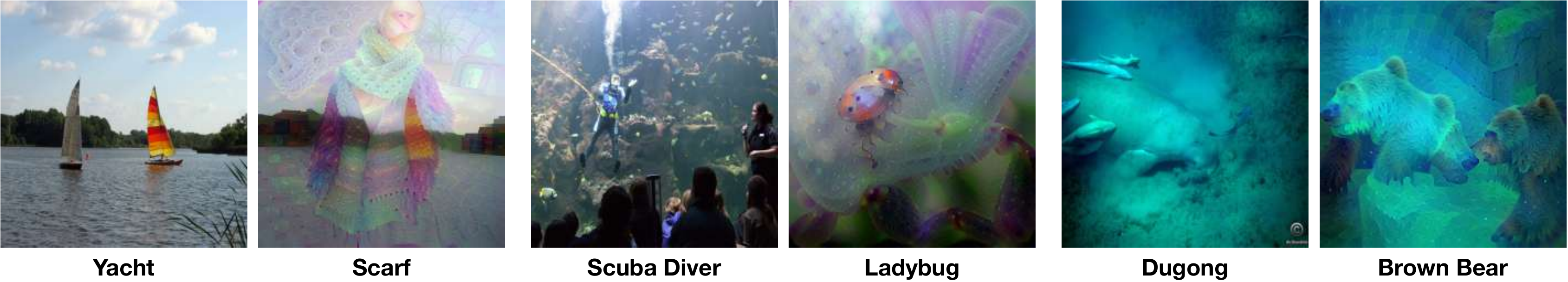}
    \label{fig:latent3}
  \end{subfigure}  
  \caption{\nj{(Top)} \nj{Generated DSVs using} soft labels: $\delta$ was set 0.6, \ie soft label $y = 0.4 y_{\text{left}} + 0.6 y_{\text{right}}$. 
  \nj{(Middle) Examples} of latent (soft-label) interpolation. \jh{(Bottom) Image Editing through latent.}}
  \label{fig:combined}
\end{figure}

\jh{To experimentally verify this, we performed a latent interpolation task and image editing, which is common \nj{in} generative models~\cite{gan1, gan2} By mixing different labels \nj{ (\(y_i = (1-\delta)y_a + \delta y_b\), where \(y_a \neq y_b\)), we generated DSVs as} depicted in Fig.~\ref{fig:combined}. 
When generating DSVs with these mixed soft labels, \nj{the generated DSVs} semantically represent the midpoint between the two classes. \nj{A generated DSV either simply contains} both images (the case of ``balloon" and ``spaniel") or semantically `fuse' objects (the case of ``lobster" and ``harp", producing an image of a harp made out of lobster claws). 
\nj{For the image editing task, we assigned the latent variable to the desired class and then aligned the image using DeepKKT loss. The result was quite surprising: the method successfully transferred the image to the desired class while maintaining the original structure. For example, the sail of a yacht was seamlessly transformed into the shape of a scarf. This task was impossible with other methods; in diffusion models, for instance, a mask would be needed to edit the image seamlessly.}
}

The fact that the generated images correctly merge the semantics of the classes suggests \nj{a couple of} significant implications: 1)  1) \jh{\textbf{New Generative Model}: This approach offers a new type of generative model as an alternative to GANs and diffusion models.} It can handle the same task without the need for training a specific model, as it leverages existing classification models for generative purposes. Furthermore, it is lightweight compared to diffusion models.
For example, \nj{as the model size of a pretrained ResNet50 for ImageNet is only one-twentieth of that for SDXL~\cite{SDXL}, DSVs show a} potential to leverage existing classification models for generative purposes. 2)  \textbf{Exploration of Classification Model Generalization}: Unlike other generative models, classification models are trained simply to predict the \nj{label} of an image. Yet, in latent interpolation \jh{and editing tasks}, they demonstrate an understanding of semantics. This implies that, despite being trained to memorize class labels, the models grasp the overall semantics of the dataset. As they can generate seemingly unseen samples by interpolation and editing.

\section{Conclusion}
\jh{\nj{In this paper, we redefined support vectors} in nonlinear deep learning models through the introduction of Deep Support Vectors (DSVs). We demonstrated the feasibility of generating DSVs using only \nj{a pretrained model}, without \nj{accessing} to the training dataset. To achieve this, we extended the KKT (Karush-Kuhn-Tucker) conditions to DeepKKT conditions \nj{and the proposed} method can be applied to any deep learning models.}

\jh{Akin to SVMs, the DeepKKT condition effectively encodes the decision boundary into DSVs. DSVs can reconstruct the model, making them useful for dataset distillation. Additionally, their visual encoding of the decision criteria can serve as a global explanation, helping to understand the model's overall behavior and decisions. Furthermore, the DeepKKT condition transforms a classification model into a generative model with high fidelity. Not only can it sample data, but it also generalizes well, allowing the use of labels as latent variables.}

\newpage
\section*{Acknowledgement}
Thank you for Hyunjin Kim, Wonhak Park, and Yeji Song for detailed discussion and feedback. This work was supported by
NRF grant (2021R1A2C3006659) and IITP grants (RS-2022-II220953, RS-2021-II211343), all funded by MSIT of the Korean
Government.

\bibliographystyle{plain}
\bibliography{neurips_2024}

\begin{thebibliography}{10}

\bibitem{svm:reconstructmulti}
Gon Buzaglo, Niv Haim, Gilad Yehudai, Gal Vardi, and Michal Irani.
\newblock Reconstructing training data from multiclass neural networks, 2023.

\bibitem{distillation_trajectory}
George Cazenavette, Tongzhou Wang, Antonio Torralba, Alexei~A. Efros, and Jun-Yan Zhu.
\newblock Dataset distillation by matching training trajectories.
\newblock In {\em Proceedings of the IEEE/CVF Conference on Computer Vision and Pattern Recognition (CVPR) Workshops}, pages 4750--4759, June 2022.

\bibitem{gan2}
Yunjey Choi, Minje Choi, Munyoung Kim, Jung-Woo Ha, Sunghun Kim, and Jaegul Choo.
\newblock Stargan: Unified generative adversarial networks for multi-domain image-to-image translation, 2018.

\bibitem{data:mnist}
Li~Deng.
\newblock The mnist database of handwritten digit images for machine learning research.
\newblock {\em IEEE Signal Processing Magazine}, 29(6):141--142, 2012.

\bibitem{convnet}
Ian~J. Goodfellow, Yoshua Bengio, and Aaron Courville.
\newblock {\em Deep Learning}.
\newblock MIT Press, Cambridge, MA, USA, 2016.
\newblock \url{http://www.deeplearningbook.org}.

\bibitem{svm:marginreconstruct}
Niv Haim, Gal Vardi, Gilad Yehudai, michal Irani, and Ohad Shamir.
\newblock Reconstructing training data from trained neural networks.
\newblock In Alice~H. Oh, Alekh Agarwal, Danielle Belgrave, and Kyunghyun Cho, editors, {\em Advances in Neural Information Processing Systems}, 2022.

\bibitem{resnet}
Kaiming He, Xiangyu Zhang, Shaoqing Ren, and Jian Sun.
\newblock Deep residual learning for image recognition.
\newblock {\em CoRR}, abs/1512.03385, 2015.

\bibitem{gan1}
Tero Karras, Samuli Laine, and Timo Aila.
\newblock A style-based generator architecture for generative adversarial networks.
\newblock {\em CoRR}, abs/1812.04948, 2018.

\bibitem{ADAM}
Diederik~P Kingma and Jimmy Ba.
\newblock Adam: A method for stochastic optimization.
\newblock {\em arXiv preprint arXiv:1412.6980}, 2014.

\bibitem{fl2}
Jakub Kone{\v{c}}n{\'y}, H.~Brendan McMahan, Daniel Ramage, and Peter Richt{\'{a}}rik.
\newblock Federated optimization: Distributed machine learning for on-device intelligence.
\newblock {\em CoRR}, abs/1610.02527, 2016.

\bibitem{dataset:cifar10}
Alex Krizhevsky.
\newblock Learning multiple layers of features from tiny images.
\newblock Technical report, 2009.

\bibitem{practicalDD}
Hyunho Lee, Junhoo Lee, and Nojun Kwak.
\newblock Practical dataset distillation based on deep support vectors, 2024.

\bibitem{distillation:nongrad1}
Songhua Liu, Kai Wang, Xingyi Yang, Jingwen Ye, and Xinchao Wang.
\newblock Dataset distillation via factorization.
\newblock In S.~Koyejo, S.~Mohamed, A.~Agarwal, D.~Belgrave, K.~Cho, and A.~Oh, editors, {\em Advances in Neural Information Processing Systems}, volume~35, pages 1100--1113. Curran Associates, Inc., 2022.

\bibitem{dataset:celebA}
Ziwei Liu, Ping Luo, Xiaogang Wang, and Xiaoou Tang.
\newblock Deep learning face attributes in the wild.
\newblock In {\em 2015 IEEE International Conference on Computer Vision (ICCV)}, pages 3730--3738, 2015.

\bibitem{svmhomo}
Kaifeng Lyu and Jian Li.
\newblock Gradient descent maximizes the margin of homogeneous neural networks.
\newblock In {\em International Conference on Learning Representations}, 2020.

\bibitem{total_prior1}
Aravindh Mahendran and Andrea Vedaldi.
\newblock Understanding deep image representations by inverting them.
\newblock {\em CoRR}, abs/1412.0035, 2014.

\bibitem{fl_start}
Brendan McMahan, Eider Moore, Daniel Ramage, Seth Hampson, and Blaise~Aguera y~Arcas.
\newblock Communication-efficient learning of deep networks from decentralized data.
\newblock In {\em Artificial intelligence and statistics}, pages 1273--1282. PMLR, 2017.

\bibitem{FL1}
H.~Brendan McMahan, Eider Moore, Daniel Ramage, and Blaise~Ag{\"{u}}era y~Arcas.
\newblock Federated learning of deep networks using model averaging.
\newblock {\em CoRR}, abs/1602.05629, 2016.

\bibitem{adversaraialuniversal}
Seyed{-}Mohsen Moosavi{-}Dezfooli, Alhussein Fawzi, Omar Fawzi, and Pascal Frossard.
\newblock Universal adversarial perturbations.
\newblock {\em CoRR}, abs/1610.08401, 2016.

\bibitem{dataset:svhn}
Yuval Netzer, Tao Wang, Adam Coates, Alessandro Bissacco, Bo~Wu, and Andrew~Y Ng.
\newblock Reading digits in natural images with unsupervised feature learning.
\newblock 2011.

\bibitem{SVMEX3}
Onuwa Okwuashi and Christopher~E. Ndehedehe.
\newblock Deep support vector machine for hyperspectral image classification.
\newblock {\em Pattern Recognition}, 103:107298, 2020.

\bibitem{pytorch}
Adam Paszke, Sam Gross, Francisco Massa, Adam Lerer, James Bradbury, Gregory Chanan, Trevor Killeen, Zeming Lin, Natalia Gimelshein, Luca Antiga, Alban Desmaison, Andreas K{\"{o}}pf, Edward~Z. Yang, Zach DeVito, Martin Raison, Alykhan Tejani, Sasank Chilamkurthy, Benoit Steiner, Lu~Fang, Junjie Bai, and Soumith Chintala.
\newblock Pytorch: An imperative style, high-performance deep learning library.
\newblock {\em CoRR}, abs/1912.01703, 2019.

\bibitem{SVMex2}
Zhiquan Qi, Bo~Wang, Yingjie Tian, and Peng Zhang.
\newblock When ensemble learning meets deep learning: a new deep support vector machine for classification.
\newblock {\em Knowledge-Based Systems}, 107:54--60, 2016.

\bibitem{CLIP}
Alec Radford, Jong~Wook Kim, Chris Hallacy, Aditya Ramesh, Gabriel Goh, Sandhini Agarwal, Girish Sastry, Amanda Askell, Pamela Mishkin, Jack Clark, Gretchen Krueger, and Ilya Sutskever.
\newblock Learning transferable visual models from natural language supervision.
\newblock {\em CoRR}, abs/2103.00020, 2021.

\bibitem{SDXL}
Robin Rombach, Andreas Blattmann, Dominik Lorenz, Patrick Esser, and Bj{\"o}rn Ommer.
\newblock High-resolution image synthesis with latent diffusion models.
\newblock In {\em Proceedings of the IEEE/CVF conference on computer vision and pattern recognition}, pages 10684--10695, 2022.

\bibitem{imagenet}
Olga Russakovsky, Jia Deng, Hao Su, Jonathan Krause, Sanjeev Satheesh, Sean Ma, Zhiheng Huang, Andrej Karpathy, Aditya Khosla, Michael~S. Bernstein, Alexander~C. Berg, and Li~Fei{-}Fei.
\newblock Imagenet large scale visual recognition challenge.
\newblock {\em CoRR}, abs/1409.0575, 2014.

\bibitem{SVMex1}
Hichem Sahbi.
\newblock Totally deep support vector machines.
\newblock {\em CoRR}, abs/1912.05864, 2019.

\bibitem{svm:theory}
Daniel Soudry, Elad Hoffer, Mor Shpigel~Nacson, Suriya Gunasekar, and Nathan Srebro.
\newblock The implicit bias of gradient descent on separable data.
\newblock In {\em International Conference on Learning Representations (ICLR)}, Vancouver, BC, Canada, 2018.
\newblock Poster presented at ICLR 2018.

\bibitem{svmpractical2}
Yichuan Tang.
\newblock Deep learning using linear support vector machines.
\newblock {\em arXiv preprint arXiv:1306.0239}, 2013.

\bibitem{svmpractical}
Jingyuan Wang, Kai Feng, and Junjie Wu.
\newblock Svm-based deep stacking networks.
\newblock In {\em Proceedings of the AAAI conference on artificial intelligence}, pages 5273--5280, 2019.

\bibitem{distillation:grad1}
Tongzhou Wang, Jun-Yan Zhu, Antonio Torralba, and Alexei~A Efros.
\newblock Dataset distillation.
\newblock {\em arXiv preprint arXiv:1811.10959}, 2018.

\bibitem{xu2019adversarial}
Han Xu, Yao Ma, Haochen Liu, Debayan Deb, Hui Liu, Jiliang Tang, and Anil~K. Jain.
\newblock Adversarial attacks and defenses in images, graphs and text: A review, 2019.

\bibitem{total_prior}
Hongxu Yin, Arun Mallya, Arash Vahdat, Jos{\'{e}}~M. {\'{A}}lvarez, Jan Kautz, and Pavlo Molchanov.
\newblock See through gradients: Image batch recovery via gradinversion.
\newblock {\em CoRR}, abs/2104.07586, 2021.

\bibitem{margin:genfclassifer}
Runpeng Yu and Xinchao Wang.
\newblock Generator born from classifier.
\newblock In {\em Thirty-seventh Conference on Neural Information Processing Systems}, 2023.

\bibitem{DSA}
Bo~Zhao and Hakan Bilen.
\newblock Dataset condensation with differentiable siamese augmentation.
\newblock {\em CoRR}, abs/2102.08259, 2021.

\bibitem{DM}
Bo~Zhao and Hakan Bilen.
\newblock Dataset condensation with distribution matching.
\newblock {\em CoRR}, abs/2110.04181, 2021.

\bibitem{DC}
Bo~Zhao, Konda~Reddy Mopuri, and Hakan Bilen.
\newblock Dataset condensation with gradient matching.
\newblock {\em CoRR}, abs/2006.05929, 2020.

\end{thebibliography}

\newpage


\appendix


\section{Societal Impacts}
\label{sec:social impace}
\nj{Our paper is closely related to Responsible AI (RAI), especially in enabling qualitative assessments of models. Our approach provides visual and intuitive explanations of a model’s decision-making criteria, offering insights that are both explanatory and responsible. Our approach of utilizing DSVs for RAI enables global explanations, surpassing traditional Explainable AI (XAI) methodologies, which usually focus on local explanations for individual inputs and cannot provide a global decision criterion. 
Furthermore, since our method is based on model inversion, it ensures safety and privacy.
While the synthesized sets in Fig.~\ref{fig:DSVSET} might appear similar to the selected sets, they do not replicate specific sample features. This is because DSVs represent a more generalized decision boundary, avoiding the inclusion of image-specific features. Consequently, DSVs enable all models using logistic loss to be more responsible.}


\begin{figure}[h]
  \centering
   \includegraphics[width=1.0\linewidth]{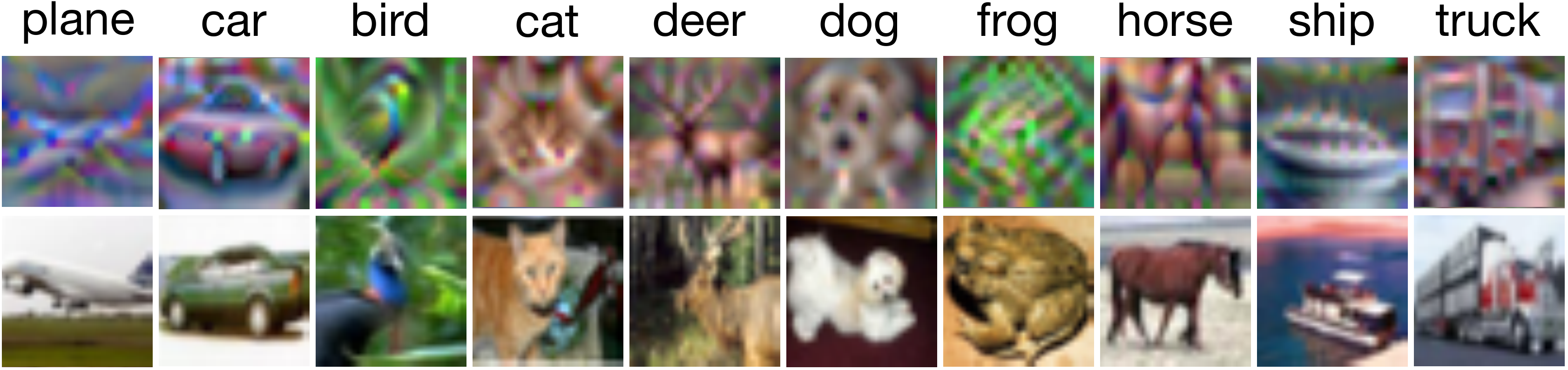}
   \caption{Comparison of synthesized images (first row) created using the DeepKKT condition initiated from noise, and selected images (second row) from the CIFAR-10 training dataset. The selected images were chosen based on \nj{$\lambda$} values, \nj{\ie each image has} the highest \nj{$\lambda$} in each class. Both synthesized and selected images demonstrate similarity at the pixel level sharing common features.}
   \label{fig:DSVSET}
\end{figure}

\section{Limitations and Future work}

\label{sec:limitations}

\nj{In this paper, we propose the DeepKKT condition, which can be applied universally to any deep models to generate deep support vectors (DSVs) that function similarly to support vectors in SVMs.} 
However, \nj{it should be noted that} the equivalence \nj{between DSVs in deep learning models and support vectors in SVMs is only described intuitively, not rigorously.} 
We have shown experimentally \nj{in Fig.~}\ref{fig:DSVSET} and intuitively \nj{in Sec.~}\ref{sec:overparam} why the DeepKKT condition should be as we suggested, but we have not derived it \nj{with rigorous math.} 
Proving this rigorously 
would be a meaningful research \nj{topic}.

\section{Intutive explanation of DeepKKT condition}
\label{sec:overparam}

\begin{figure}[t]
  \centering
   \includegraphics[width=0.9\linewidth]{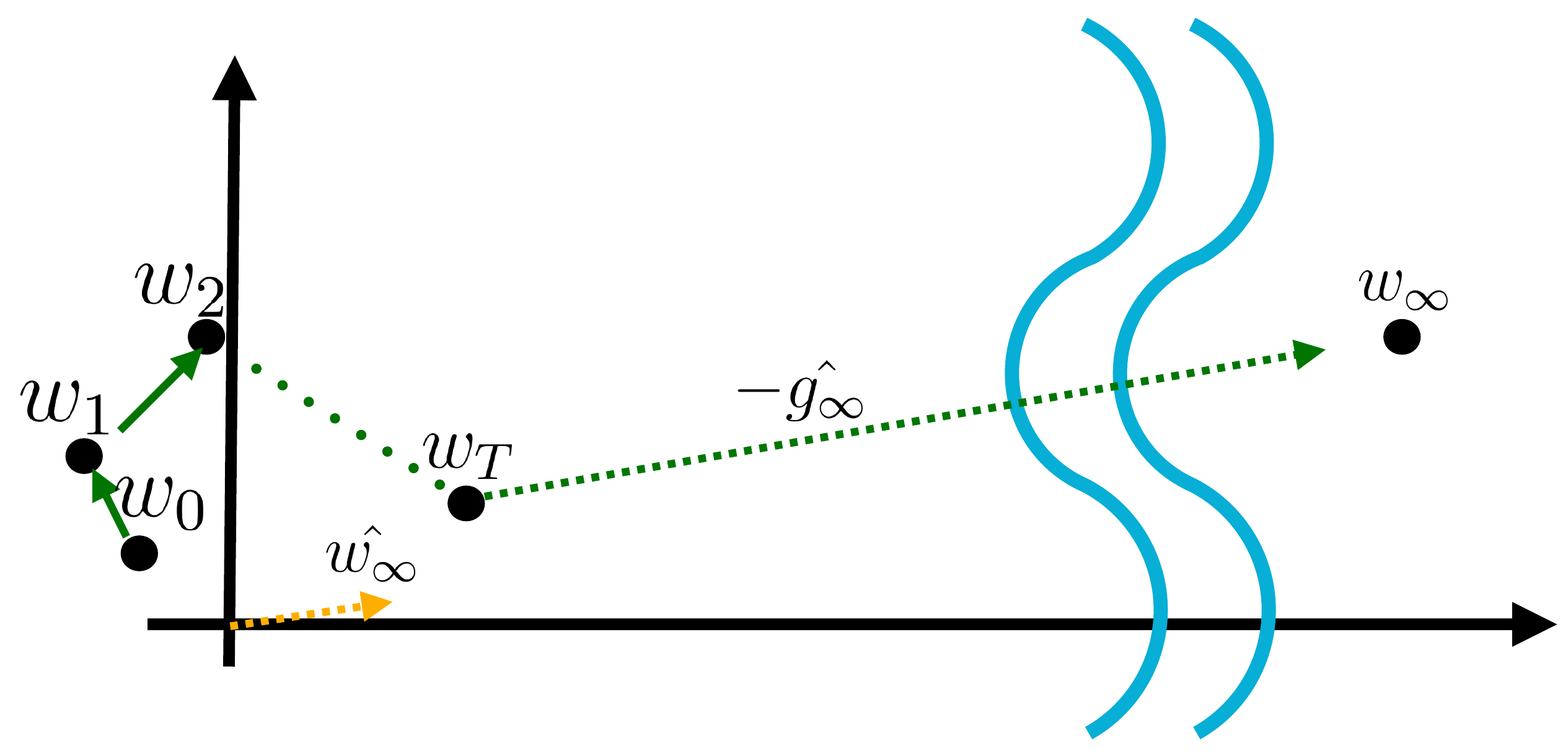}
   \caption{\jh{The stationarity condition with a logistic loss. Even though the direction of the gradient $\hat{g}$ converges, the size of the gradient does not go to zero. Therefore, the direction of the converged gradient weight $\hat{w_\infty}$ aligns with $\hat{g}$.}}
   \label{fig:intuitiveun}
\end{figure}

\nj{In DeepKKT, many conditions make sense, except for one. For instance, the primal feasibility condition and the manifold condition are reasonable, and the dual feasibility condition can be regarded as importance sampling. However, the most counterintuitive part is the stationarity condition:}
\begin{equation}
\hh{L_{\text{stat}} = D(\theta^*, -\sum_{i=1}^n \lambda_i \nabla_\theta L(\Phi(x_i;\theta^*), y_i))}
\end{equation}

\nj{In this section, we will explain the dynamics of DSVs in an overparameterized deep network and how it is connected to deep learning. Below is a quick analogy of \cite{svm:theory} to illustrate this connection.}

A deep learning \nj{model} follows \nj{the} following ODE:

\begin{equation}
    w_{t+1} = w_t - \eta \nabla L(x,y;w_t).
\end{equation}


\jh{\nj{Here,} $\eta$ is the learning rate and $t$ is the optimization step. \nj{The loss} $L$ does not go to zero since deep learning models usually exploit a \nj{loss} function with a logistic tail, such as the cross-entropy \nj{loss}, and the gradient of the least confident sample (support vector) dominates overall gradient. Thus, there exists a convergence of the gradient direction $g_\infty \coloneqq \hat{\nabla} L$. There also exists a time $T$ where the gradient direction converges to $g_\infty - \varepsilon$ for a sufficiently small $\varepsilon$. As illustrated in Fig.~\ref{fig:intuitiveun}, $w$ moves toward the direction of $-g_\infty$. Therefore, $\hat{w}_\infty \approx -g_\infty$.}

\jh{This is \nj{for what} stationarity condition wants to seek. The direction of $g_\infty$, by using only \nj{a} few support vectors.}

\section{Implementation Details}
\label{sec:detail}

To obtain the results in Table~\ref{table:fewshot} and Fig.~\ref{fig:transfer}, the ConvNet architecture~\cite{convnet} was used for pretraining \(\Phi(\cdot; \theta)\) on the \nj{SVHN} dataset~\cite{dataset:svhn},} a digit dataset with dimensions similar to CIFAR-10~\cite{dataset:cifar10}. \jh{For ImageNet, we used \nj{the} ResNet50 model~\cite{resnet} with \nj{the} original setting \nj{in the paper}. Specifically, we used \nj{the} pretrained model in torchvision library in pytorch~\cite{pytorch}. \jh{For visualizing synthesized DSVs in ImageNet, we increased the contrast in 224x224 dimensions.} When calculating $L_\text{stat}$, we averaged the distance per parameter. \nj{In Alg.~\ref{alg:KKT}, $\eta$ was set to 5.}}

\jh{To synthesize DSVs in ImageNet, we used translation, crop, cutout, flip, and noise for augmentation, with hyperparameters set to 0.125, 0.2, 0.15, 0.5, and 0.01, respectively. In Eq.~(\ref{eq:main_equation}), we set $\alpha$ to 2e-5, $\beta$ to 40, and $\gamma$ to 1e-6. When calculating $L_\text{stationarity}$, we averaged the distance per parameter.}

\jh{For dataset distillation in Table~\ref{table:fewshot}, we used translation, crop, flip, and noise for augmentation, with hyperparameters set to 0.125, 0.2, and 0.5, respectively. In Eq.~(\ref{eq:main_equation}), we set $\alpha$ to 2e-3, and both $\beta$ and $\gamma$ to 0. For retraining models with synthesized images, we used a learning rate of 1e-4 while \nj{the other parameters set to the default values of} the Adam optimizer~\cite{ADAM}.}

\nj{To obtain the pretrained weight $\theta^*$ for CIFAR10 and CIFAR100,} we chose the ConvNet architecture~\cite{convnet}, a common choice in deep learning. This architecture includes sequential convolutional layers followed by max pooling, and a single fully-connected layer for classification. The learning rate was set to \nj{\(10^{-3}\)} with a weight decay of \nj{\(0.005\)} using the Adam optimizer. Additionally, we employed flipping and cropping techniques, with settings differing from those used for DSVs reconstruction to ensure fair comparison. For pretraining \(\Phi\) on the Street View House Numbers (SVHN) dataset~\cite{dataset:svhn}, a digit dataset with dimensions similar to CIFAR-10~\cite{dataset:cifar10}, we exclusively trained the fully-connected layer of the CIFAR-10 pre-trained ConvNet. This approach resulted in a training accuracy of 80\%. 

\section{DSVs \nj{by} Selection}
\jh{Fig.~\ref{fig:entropyhigh} shows the selected images with large Lagrangian multipliers $\lambda$'s, which correspond to the candidates used in Fig.~\ref{fig:correlation}. Surprisingly, there is a meaningful match between the selected DSVs and the synthesized DSVs in the CIFAR-10 dataset, as shown in Fig.~\ref{fig:DSVSET}. This implies that synthesizing DSVs corresponds to \nj{reviving} training data that lie on the boundary manifolds.
}
\begin{figure}[h]
  \centering
   \includegraphics[width=0.9\linewidth]{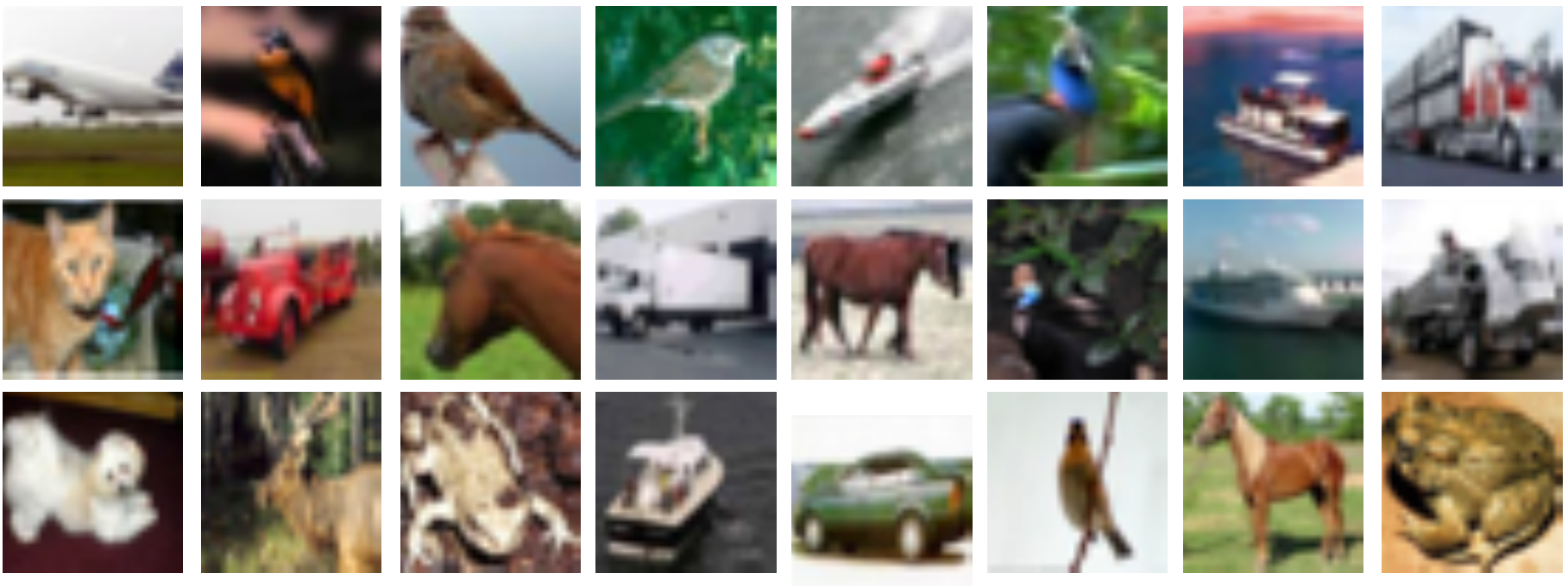}
   \caption{Images of DSV candidates (Selected in the CIFAR-10 dataset).}
   \label{fig:entropyhigh}
\end{figure}

\begin{figure}[h]
  \centering
   \includegraphics[width=0.5\linewidth]{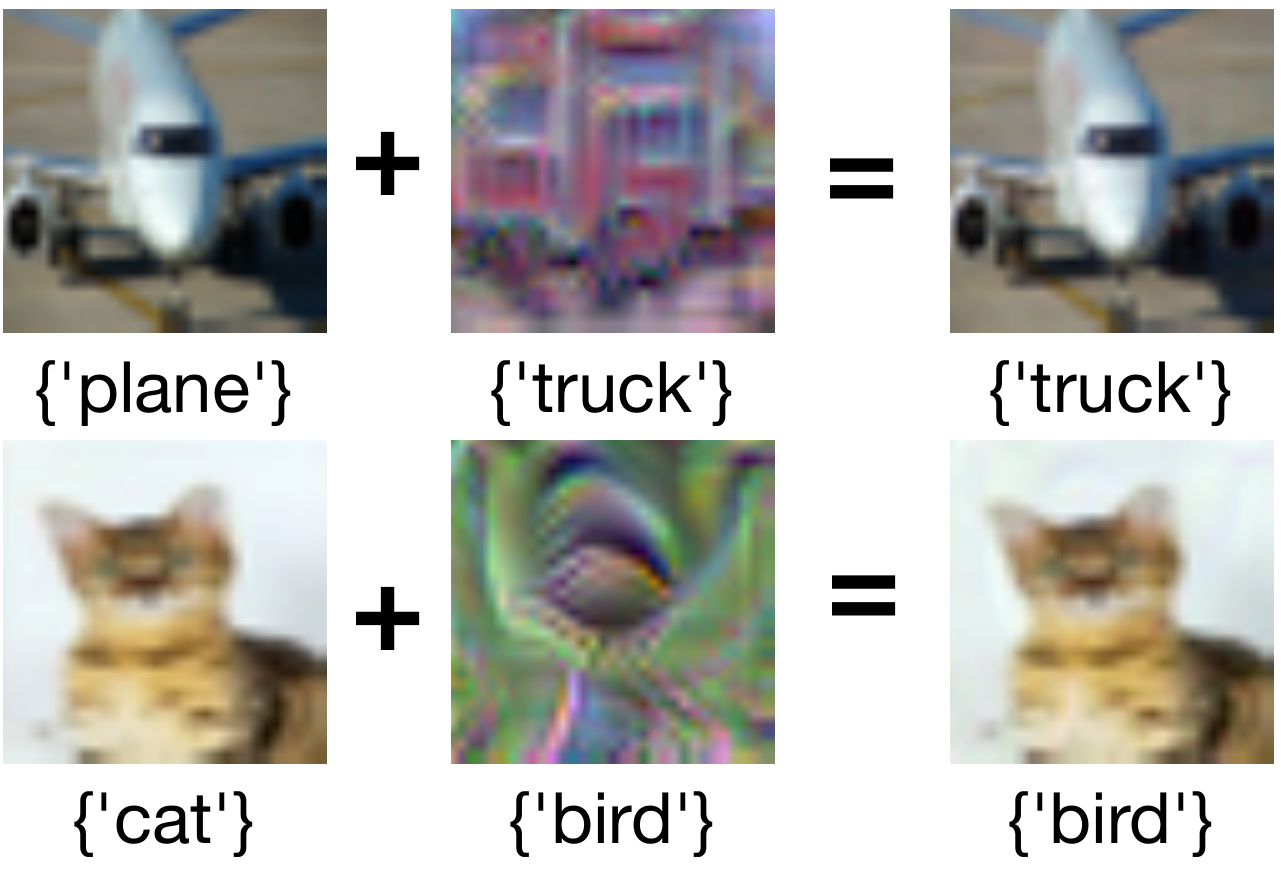}
   \caption{\gm{Examples of adversarial attack in the CIFAR-10 dataset}}
   \label{fig:adversarial}
\end{figure}

\begin{figure}[h]
  \centering
   \includegraphics[width=0.9\linewidth]{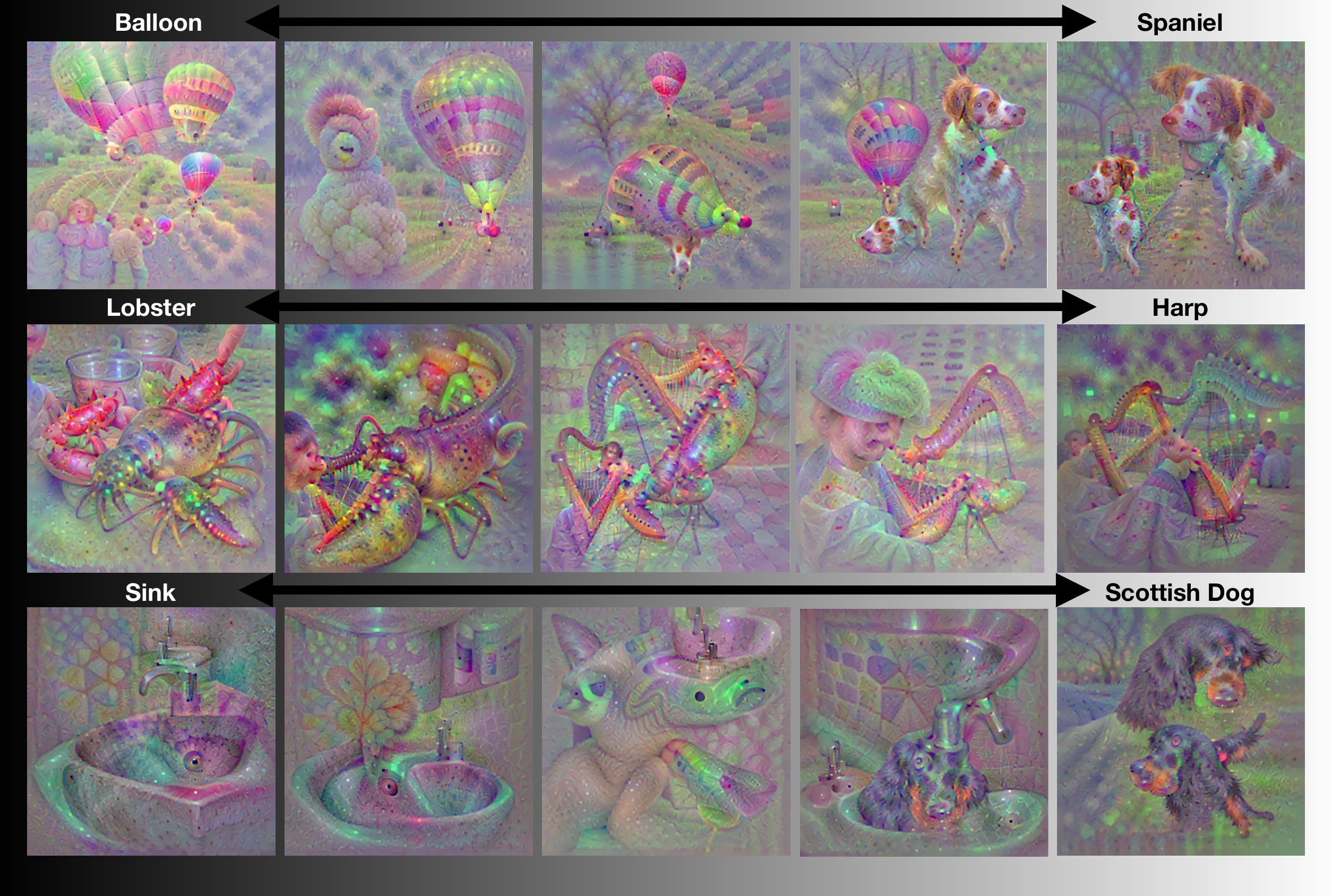}
   \caption{More examples of latent interpolation in the ImageNet dataset}
   \label{fig:latentmore}
\end{figure}

\begin{figure}[h]
  \centering
   \includegraphics[width=0.9\linewidth]{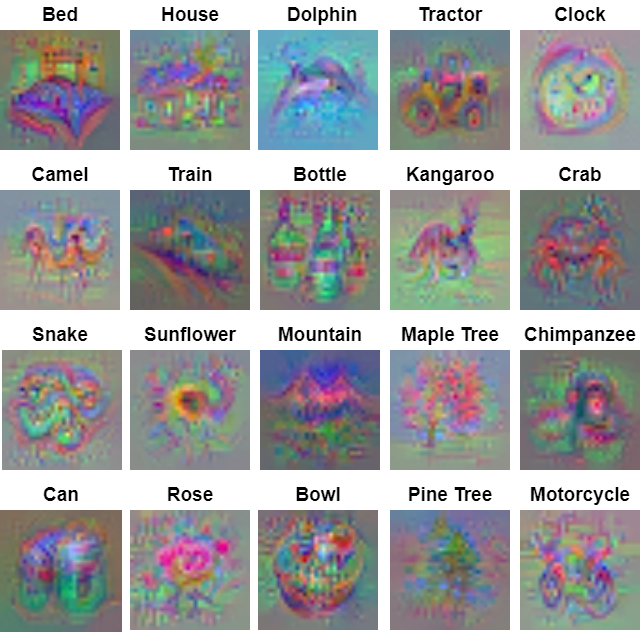}
   \caption{\gm{More examples of generated images with CIFAR-100 dataset}}
   \label{fig:dsv_cifar100}
\end{figure}

\begin{figure}[h]
  \centering
   \includegraphics[width=0.9\linewidth]{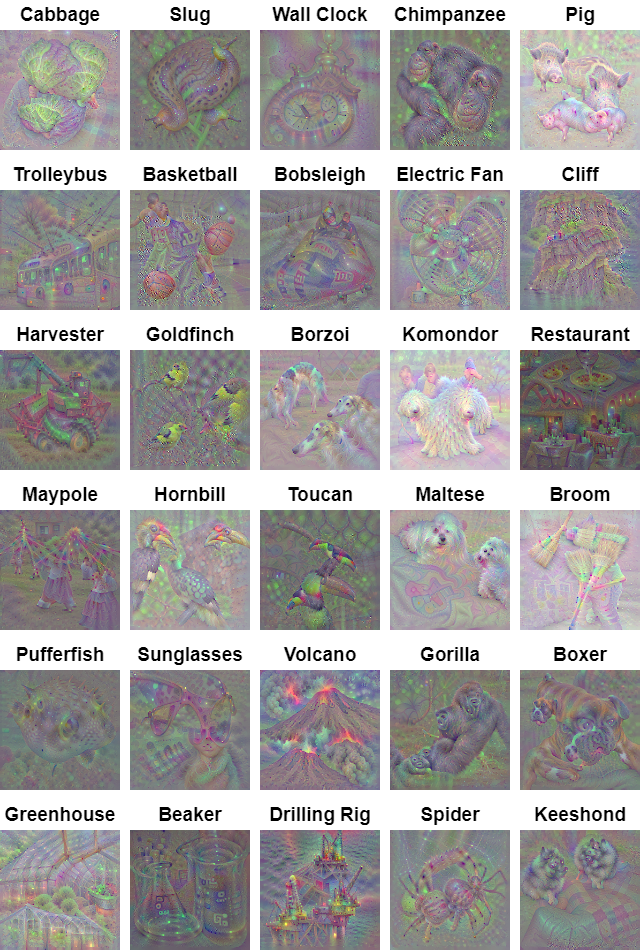}
   \caption{More examples of generated images with ImageNet dataset}
   \label{fig:dsv_imagenet}
\end{figure}

\begin{figure}[h]
  \centering
   \includegraphics[width=0.9\linewidth]{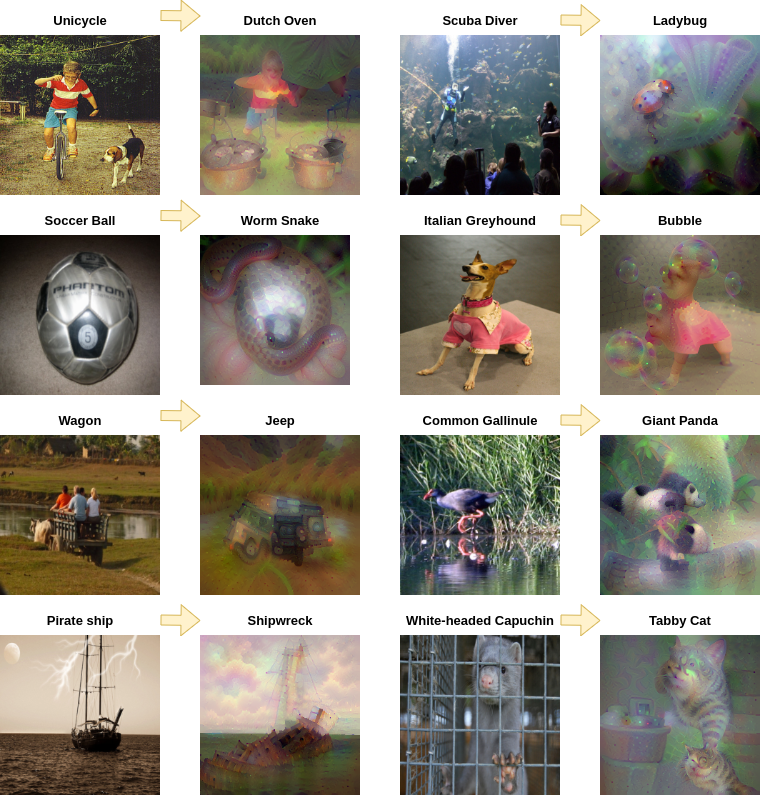}
   \caption{\gm{More examples of image editing. The images to the left of the arrows represent the initial images before training, while those to the right depict the edited images after training.}}
   \label{fig:dsv_editing}
\end{figure}

\begin{figure}[h]
  \centering
   \includegraphics[width=0.9\linewidth]{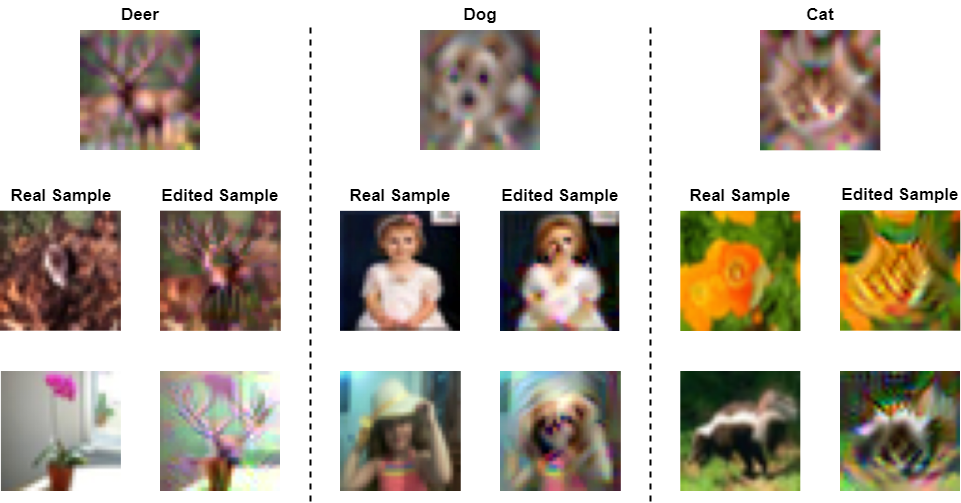}
   \caption{\gm{More examples of image editing. The images to the left of the arrows represent the initial images before training, while those to the right depict the edited images after training.}}
   \label{fig:decision_criterion}
\end{figure}

\section{More Characteristics of DSVs}
\paragraph{DSVs Are Full of Discriminative Information}

\jh{In Fig.~\ref{fig:adversarial}, we conducted an experiment by mixing a randomly sampled image from the real dataset with an image from the DSVs. 
Upon observation, the mixed image is virtually indistinguishable from an image obtained solely from the real dataset.}
\jh{It is noteworthy to highlight this situation resembles that of an adversarial attack \citep{xu2019adversarial, adversaraialuniversal}, yet we did not apply gradient descent to the image; we simply mixed two images. This suggests that the \nj{discriminative} informational density in a single DSV image is substantially greater than that in a randomly sampled image. The fact that the DSV's characteristics remained dominant in the classification, 
underscores the significant role of DSVs in explaining the model's classification ability.}

\section{More Examples}
\jh{In Fig.~\ref{fig:latentmore}, examples of latent interpolations between target labels are presented. The smoothness of these interpolations within the latent space indicates that the semantic information learned from the training data has been effectively applied during the DSV generation process. This observation provides evidence that the DeepKKT optimization successfully conducts the generative process.}

\gm{Fig.~\ref{fig:dsv_cifar100} and ~\ref{fig:dsv_imagenet} provide examples of deep support vectors generated using the CIFAR-100 and ImageNet datasets, respectively. Fig.~\ref{fig:dsv_editing} presents additional examples related to image editing.}


\jh{Fig.~\ref{fig:decision_criterion} empirically supports on our assertion on decision criterion. Starting from CIFAR100 random images and CIFAR10-pretrained models, we edited the image \hh{CIFAR10} labels as latents. The edited images changes the image following decision criterions in generated DSVs. 1) For editing images to deer, antler grows. 2) For dog editing, facial dots are generated. 3) For cat editing, pointed traiangler features are generated.}

\section{Algorithm}
\label{sec:alg}
Alg.~\ref{alg:KKT} presents our algorithm of generating Deep Support Vectors (DSVs).
Initialized either from a noise $x_i^s \sim \mathcal{N}(0,I)$ or a real sample, it iterates to obtain the primal $X^S$ and dual $\Lambda^S$ variables.  

\begin{algorithm*}[h]
\caption{Support Vector Refinement for Deep Learning Model }\label{alg:KKT}
\begin{algorithmic}[1]
\Require Pretrained classifier $\Phi(\cdot; \theta)$, loss function $L$, augmentation function set $A$, number of DSV candidate $N$, number of class  $C$, hyperparameters $\alpha$, $\beta$
\Ensure Freeze classifier $\Phi(\cdot; \theta)$

\State Initialize $N \times C$ number of support vector candidates
\For{$i = 1$ to $C$}
    \State sample $N$ number of $(x^s_i, \lambda^s_i)$ for label $y^s_i = i$
\EndFor
\State Define $X^S = \{ x^s_i \mid i \in [C], s \in [N]\}$
\State Define $\Lambda^S = \{ \lambda^s_i \mid i \in [C], s \in [N]\}$
\Repeat
    \State $L_{\text{primal}}(X^S) = \sum_{s=1}^{N} \sum_{i=1}^{C} L(\Phi(x^s_i;\theta), y^s_i)$
    \State $L_{\text{stationary}}(X^S) = \|\theta + \sum_{s=1}^{N} \sum_{i=1}^{C} \lambda^s_i y^s_i \nabla_\theta \Phi(x^s_i;\theta)\|^2_2$
    \State $L_{\text{kkt}}(X^S) = \beta_1 \cdot L_{\text{primal}}(X^S) +  L_{\text{stationary}}(X^S)$
    \State $L_{\text{prior}}$ = $\beta_2 \cdot L_{\text{tot}}(X) + \beta_3 L_{\text{norm}}(X)$
    \State Sample $f_A \in A$
    \State Define $AX^S = \{ f_A(x^s_i) \mid x^s_i \in X^S\}$
    \State $L_{\text{akkt}}(X^S) = L_{\text{kkt}}(AX^S)$  
    \State $L_{\text{total}}(X^S) = L_{\text{kkt}}(X^S) + \eta \cdot L_{\text{akkt}}(X^S)$ + \jh{$L_{\text{prior}}$}
    \State Update $X^S \leftarrow X^S + \nabla_{X^S} L_{\text{total}}(X^S)$
    \State Update $\Lambda^S \leftarrow \Lambda^S + \nabla_{\Lambda^S} L_{\text{total}}(X^S)$
    \State Remove $x^s_i$s for corresponding $\lambda^s_i < 0$
\Until{$X^S$ converges}
\State \Return Set of DSV : $X^S$
\end{algorithmic}
\end{algorithm*}

\newpage
\clearpage
\section*{NeurIPS Paper Checklist}

The checklist is designed to encourage best practices for responsible machine learning research, addressing issues of reproducibility, transparency, research ethics, and societal impact. Do not remove the checklist: {\bf The papers not including the checklist will be desk rejected.} The checklist should follow the references and follow the (optional) supplemental material.  The checklist does NOT count towards the page
limit. 

Please read the checklist guidelines carefully for information on how to answer these questions. For each question in the checklist:
\begin{itemize}
    \item You should answer \answerYes{}, \answerNo{}, or \answerNA{}.
    \item \answerNA{} means either that the question is Not Applicable for that particular paper or the relevant information is Not Available.
    \item Please provide a short (1–2 sentence) justification right after your answer (even for NA). 
\end{itemize}

{\bf The checklist answers are an integral part of your paper submission.} They are visible to the reviewers, area chairs, senior area chairs, and ethics reviewers. You will be asked to also include it (after eventual revisions) with the final version of your paper, and its final version will be published with the paper.

The reviewers of your paper will be asked to use the checklist as one of the factors in their evaluation. While "\answerYes{}" is generally preferable to "\answerNo{}", it is perfectly acceptable to answer "\answerNo{}" provided a proper justification is given (e.g., "error bars are not reported because it would be too computationally expensive" or "we were unable to find the license for the dataset we used"). In general, answering "\answerNo{}" or "\answerNA{}" is not grounds for rejection. While the questions are phrased in a binary way, we acknowledge that the true answer is often more nuanced, so please just use your best judgment and write a justification to elaborate. All supporting evidence can appear either in the main paper or the supplemental material, provided in appendix. If you answer \answerYes{} to a question, in the justification please point to the section(s) where related material for the question can be found.

IMPORTANT, please:
\begin{itemize}
    \item {\bf Delete this instruction block, but keep the section heading ``NeurIPS paper checklist"},
    \item  {\bf Keep the checklist subsection headings, questions/answers and guidelines below.}
    \item {\bf Do not modify the questions and only use the provided macros for your answers}.
\end{itemize}


\begin{enumerate}

\item {\bf Claims}
    \item[] Question: Do the main claims made in the abstract and introduction accurately reflect the paper's contributions and scope?
    \item[] Answer: \answerYes{} 
    \item[] Justification: The abstract and introduction provide a clear overview of the paper’s primary contributions, particularly highlighting the introduction of the DeepKKT condition and the concept of Deep Support Vectors (DSVs). These contributions are accurately reflected in the body of the paper through theoretical formulations, experimental evidence, and practical applications of DSVs, aligning with the claims of the paper.


\item {\bf Limitations}
    \item[] Question: Does the paper discuss the limitations of the work performed by the authors?
    \item[] Answer: \answerYes{} 
    \item[] Justification: See limitation section in Sec.~\ref{sec:limitations} The limitations are presented in Section \textbackslash{}ref\{sec:limitations\}. The paper notes that the DeepKKT condition’s equivalence to traditional support vector representations remains intuitive rather than rigorously proven. It also acknowledges the computational challenges of enforcing all KKT conditions explicitly, especially in high-dimensional deep learning contexts.


\item {\bf Theory Assumptions and Proofs}
    \item[] Question: For each theoretical result, does the paper provide the full set of assumptions and a complete (and correct) proof?
    \item[] Answer: \answerYes{} 
    \item[] Justification: The theoretical framework is grounded in adapted KKT conditions, with assumptions explicitly stated for applying these in high-dimensional, multi-class deep learning settings. Although some proofs are briefly outlined, the core theoretical justifications are complete, with further details in the appendix for additional clarification.


    \item {\bf Experimental Result Reproducibility}
    \item[] Question: Does the paper fully disclose all the information needed to reproduce the main experimental results of the paper to the extent that it affects the main claims and/or conclusions of the paper (regardless of whether the code and data are provided or not)?
    \item[] Answer: \answerYes{} 
    \item[] Justification: The paper provides comprehensive details on the model architectures, datasets, augmentation strategies, and hyperparameters used in the experiments. Key implementation choices, such as optimizer configurations and specific data augmentation techniques, are described in the supplementary materials.


\item {\bf Open access to data and code}
    \item[] Question: Does the paper provide open access to the data and code, with sufficient instructions to faithfully reproduce the main experimental results, as described in supplemental material?
    \item[] Answer: \answerYes{} 
    \item[] Justification:  We submitted code in supplementary.

\item {\bf Experimental Setting/Details}
    \item[] Question: Does the paper specify all the training and test details (e.g., data splits, hyperparameters, how they were chosen, type of optimizer, etc.) necessary to understand the results?
    \item[] Answer: \answerYes{} 
    \item[] Justification: The experimental setup includes clear specifications on data splits, selected hyperparameters, and optimizers for each experimental task. Additional settings, such as augmentation parameters and model architecture details, are included, enabling a comprehensive understanding of the experimental environment.


\item {\bf Experiment Statistical Significance}
    \item[] Question: Does the paper report error bars suitably and correctly defined or other appropriate information about the statistical significance of the experiments?
    \item[] Answer: \answerYes{} 
    \item[] Justification: The paper includes tables with error bars representing the variability of DSV performance on benchmarks. The error bars are correctly computed, taking into account variations across training runs, which supports the robustness of the claims made.


\item {\bf Experiments Compute Resources}
    \item[] Question: For each experiment, does the paper provide sufficient information on the computer resources (type of compute workers, memory, time of execution) needed to reproduce the experiments?
    \item[] Answer: \answerYes{} 
    \item[] Justification: The paper specifies the use of GPUs for all major experiments and provides approximate training times. Resources are sufficiently detailed to allow for replication, indicating required compute types and time estimates for reproducibility.

    
\item {\bf Code Of Ethics}
    \item[] Question: Does the research conducted in the paper conform, in every respect, with the NeurIPS Code of Ethics \url{https://neurips.cc/public/EthicsGuidelines}?
    \item[] Answer: \answerYes{} 
    \item[] Justification: he work aligns with the NeurIPS Code of Ethics, emphasizing privacy and responsible AI principles, especially concerning the responsible use of DeepKKT for dataset distillation and model inversion, without accessing sensitive data.

\item {\bf Broader Impacts}
    \item[] Question: Does the paper discuss both potential positive societal impacts and negative societal impacts of the work performed?
    \item[] Answer: \answerYes{} 
    \item[] Justification: The paper discusses societal impacts, especially the benefits of interpretability in AI models and potential concerns about model inversion’s misuse. It acknowledges the ethical considerations associated with generating data from sensitive models, encouraging responsible handling.

    
\item {\bf Safeguards}
    \item[] Question: Does the paper describe safeguards that have been put in place for responsible release of data or models that have a high risk for misuse (e.g., pretrained language models, image generators, or scraped datasets)?
    \item[] Answer: \answerNA{} 
    \item[] Justification:  The paper does not release high-risk models or data that require specific safeguards.

\item {\bf Licenses for existing assets}
    \item[] Question: Are the creators or original owners of assets (e.g., code, data, models), used in the paper, properly credited and are the license and terms of use explicitly mentioned and properly respected?
    \item[] Answer: \answerYes{} 
    \item[] Justification: The paper appropriately references all datasets (e.g., CIFAR-10, ImageNet) and model architectures, following proper citation practices.


\item {\bf New Assets}
    \item[] Question: Are new assets introduced in the paper well documented and is the documentation provided alongside the assets?
    \item[] Answer: \answerNA{} 
    \item[] Justification: The paper does not introduce any new assets, such as unique datasets or models, that require documentation.


\item {\bf Crowdsourcing and Research with Human Subjects}
    \item[] Question: For crowdsourcing experiments and research with human subjects, does the paper include the full text of instructions given to participants and screenshots, if applicable, as well as details about compensation (if any)? 
    \item[] Answer: \answerNA{} 
    \item[] Justification: The paper does not involve human subjects or crowdsourcing experiments.


\item {\bf Institutional Review Board (IRB) Approvals or Equivalent for Research with Human Subjects}
    \item[] Question: Does the paper describe potential risks incurred by study participants, whether such risks were disclosed to the subjects, and whether Institutional Review Board (IRB) approvals (or an equivalent approval/review based on the requirements of your country or institution) were obtained?
    \item[] Answer: \answerNA{} 
    \item[] Justification: The paper does not involve human subjects research, so IRB approval is not applicable.


\end{enumerate}

\end{document}